\documentclass[10pt,twocolumn,letterpaper]{article}

\usepackage{cvpr}              

\usepackage{times}
\usepackage{epsfig}
\usepackage{amsmath}
\usepackage{amssymb}
\usepackage{booktabs}

\usepackage{comment}
\usepackage{color}
\usepackage{epsfig}
\usepackage{tabularx}
\usepackage{booktabs}
\usepackage{multirow}
\usepackage{bm}
\usepackage{algorithm,algpseudocode}
\usepackage[table,xcdraw,dvipsnames]{xcolor}
\usepackage{colortbl}
\usepackage{breakcites}
\usepackage{dsfont}
\usepackage{bbm}
\usepackage{makecell}
\usepackage{marvosym}

\usepackage[dvipsnames]{xcolor}

\definecolor{cvprblue}{rgb}{0.21,0.49,0.74}
\usepackage[pagebackref,breaklinks,colorlinks,citecolor=cvprblue]{hyperref}

\definecolor{myblue}{RGB}{23,183,241}
\definecolor{mygray}{RGB}{230,230,230}
\definecolor{myblue2}{RGB}{198,213,250}
\definecolor{mygreen}{RGB}{197,224,180}
\definecolor{myyellow}{RGB}{255,230,153}
\definecolor{myorange}{RGB}{244,171,131}

\def\paperID{1398} 
\def\confName{CVPR}
\def\confYear{2024}

\newcommand{\xjqi}[1]{\textcolor{magenta}{{[\textbf{xjqi}: #1]}}}

\title{RegionPLC: Regional Point-Language Contrastive Learning for Open-World 3D Scene Understanding}

\vspace{-0.3cm}
\author{%
  Jihan Yang\textsuperscript{1}\footnotemark[1] \quad
  Runyu Ding\textsuperscript{1}$^*$ \quad
  Weipeng Deng\textsuperscript{1} \quad
  Zhe Wang\textsuperscript{2} \quad
  Xiaojuan Qi\textsuperscript{1} \vspace{.2em}\\
  \textsuperscript{1}The University of Hong Kong \quad  \textsuperscript{2}SenseTime Research \\
  {\tt\small \url{https://jihanyang.github.io/projects/RegionPLC}}
}

\begin{document}
\maketitle
{\fnsymbol{footnote}}
\footnotetext[1]{Equal contribution: \{jhyang, ryding\}@eee.hku.hk}

\begin{abstract}
We propose a lightweight and scalable \textbf{Region}al \textbf{P}oint-\textbf{L}anguage \textbf{C}ontrastive learning framework, namely \textbf{RegionPLC}, for open-world 3D scene understanding, aiming to identify and recognize open-set objects and categories. 
Specifically, based on our empirical studies, we introduce a 3D-aware SFusion strategy that fuses 3D vision-language pairs derived from multiple 2D foundation models, yielding high-quality, dense region-level language descriptions without human 3D annotations. Subsequently, we devise a region-aware point-discriminative contrastive learning objective to enable robust and effective 3D learning from dense regional language supervision. We carry out extensive experiments on ScanNet, ScanNet200, and nuScenes datasets, and our model outperforms prior 3D open-world scene understanding approaches by an average of 17.2\% and 9.1\% for semantic and instance segmentation, respectively, while maintaining greater scalability and lower resource demands. Furthermore, our method has the flexibility to be effortlessly integrated with language models to enable open-ended grounded 3D reasoning without extra task-specific training. Code is available at \href{https://github.com/CVMI-Lab/PLA}{github}. 
\end{abstract}

\section{Introduction}
\vspace{-0.2cm}




Open-world 3D scene understanding aims to equip models with the ability to accurately perceive and identify open-set objects and categories from 3D data, such as point clouds. This ability is crucial for real-world applications where objects from open-set categories are prevalent  ~\cite{bojarski2016end, zeng2018learning}. 
However, this task poses significant challenges due to the scarcity of dense 3D semantic annotations, which are difficult to gather and scale to a large vocabulary space.

Fortunately, the abundance of paired image and text data from the Internet, featuring a vast semantic vocabulary, has enabled 2D vision-language models to exhibit exceptional open-world image comprehension capabilities. These abilities span various tasks, such as image captioning \cite{wang2022ofa, alayrac2022flamingo}, grounding \cite{wu2022grit, peng2023kosmos2}, and dense semantic prediction \cite{li2022languagedriven, zhou2022detecting, li2022grounded}. Consequently, recent research has been inspired to leverage these models to generate pseudo supervision such as dense semantic features~\cite{peng2022openscene, zeng2023clip2} and language descriptions~\cite{ding2022language, ding2023lowis3d} for training 3D models, thereby enabling open-world inference without relying on image modalities.

Despite advancements, existing solutions still exhibit limitations.
For instance, feature distillation-based methods \cite{peng2022openscene, zeng2023clip2}-- despite harvesting dense supervision-- suffer from the constraints of 2D feature qualities and 
require resource-intensive feature extraction, fusion, and storage processes, preventing them from being scaled up with more advanced 3D architectures and larger 3D datasets.
 Additionally, while \cite{ding2022language} utilizes pseudo 3D-language pairs to enable direct learning from large-vocabulary language supervision, it suffers from sparse supervision provided by image captioning models. Considering the recent success of 2D foundation models in image- and region-level vision-language learning, we explore combining their strengths to enrich vocabulary and construct high-quality region-level 3D-language associations. By doing so, our method can yield denser 3D-language supervision and circumvent the knowledge limitations of a single foundation model, facilitating resource-efficient and large-vocabulary 3D learning.

To this end, we propose a holistic \textbf{Region}al \textbf{P}oint \textbf{L}anguage \textbf{C}ontrastive learning framework, named \textbf{RegionPLC}. This framework generates and fuses diverse region-level captions from powerful 2D vision-language models, which are subsequently mapped to 3D for constructing region-level 3D and language pairs. These paired data are then incorporated into a region-aware point-discriminative contrastive learning framework, enabling 3D open-world learning from dense language supervision.

Specifically, we begin by conducting a comprehensive examination of various 2D foundation models (e.g., image captioning \cite{wang2022ofa}, dense captioning \cite{wu2022grit, peng2023kosmos2}, and detection models \cite{zhou2022detecting}) along with visual prompting techniques for their capability to generate region-level 3D-language pairs.
Based on our examination, we propose a supplementary-oriented fusion strategy that leverages the geometric relationship of regions in 3D space to alleviate ambiguities and conflicts encountered when combining paired 3D-language data from multiple 2D models, ultimately delivering high-quality dense region-level 3D-language pairs.
Furthermore, with region-level language data, we introduce a region-aware point-discriminative contrastive loss that prevents the optimization of point-wise embeddings from being disturbed by nearby points from unrelated semantic categories, enhancing the discriminativeness of learned point-wise embeddings. The region-aware design further normalizes the contribution of multiple region-level 3D-language pairs, regardless of their region sizes, making feature learning more robust. 
Finally, by harvesting the 3D-language associations, our RegionPLC can be effortlessly integrated with language models to enable open-ended 3D reasoning with grounding abilities without requiring task-specific data for training.

We conduct extensive experiments on ScanNet \cite{dai2017scannet}, ScanNet200 \cite{rozenberszki2022language}, and nuScenes \cite{caesar2020nuscenes} datasets, covering both 3D indoor and outdoor scenarios. Our method significantly outperforms existing open-world scene understanding methods, achieving an average of $17.2\%$ gains in terms of unseen category mIoU for semantic segmentation and an average of $9.1\%$ gains in terms of unseen category mAP$_{50}$ for instance segmentation.
RegionPLC demonstrates promising zero-shot segmentation performance, attaining $40.5\%$ and $1.8\%$ higher foreground mIoU compared to PLA \cite{ding2022language} and OpenScene \cite{peng2022openscene}, respectively. Notably, it achieves this performance while consuming only $17\%$ of OpenScene's \cite{peng2022openscene} training cost and $5\%$ of its storage requirements.
Furthermore, RegionPLC can also be combined with OpenScene to deliver $5.8\%$ and $10.0\%$ gains in foreground mIoU and mAcc, respectively. 

\section{Related Work}
\vspace{-0.2cm}

\noindent\textbf{3D Scene Understanding.} 
3D semantic and instance segmentation are two fundamental tasks for scene understanding, which predict each point's semantic meaning (and instance IDs) in a  3D point cloud.
For semantic feature extraction and prediction, existing approaches design customized point convolutions applied on raw point clouds~\cite{wu2019pointconv,thomas2019KPConv,xu2021paconv} or employ sparse convolution~\cite{graham2017submanifold} to develop voxel-based networks~\cite{graham20183d,choy20194d} or transformers~\cite{lai2022stratified} based on 3D grids. For instance-level prediction, representative approaches often use a bottom-up strategy that groups points to form object proposals~\cite{jiang2020pointgroup,vu2022softgroup,vu2022softgroup++}, or first predicts 3D bounding boxes and then refines the object masks using a top-down solution~\cite{yi2019gspn,yang2019learning,kolodiazhnyi2023top}. 
Though achieving outstanding results on close-set benchmark datasets, they always struggle with open-world recognition.

\noindent\textbf{Open-world 3D Understanding.} 
Open-world 3D understanding~\cite{zhang2023clip,liu2023openshape,ding2022language,peng2022openscene}
aims to recognize novel categories that are unseen during training.
Most recently, the high open-world capability of 2D foundation models ~\cite{radford2021learning,alayrac2022flamingo} trained on massive multi-modality data has inspired recent approaches to leverage them for 3D open-world understanding. 
One line of work~\cite{zhang2022pointclip,huang2022clip2point,peng2022openscene,jatavallabhula2023conceptfusion,huang2023openins3d,takmaz2023openmask3d} focuses on incorporating these 2D foundation models in the \textit{inference stage for open-world recognition}, which mainly conducts open-world semantic prediction on the image modality using vision-language models~\cite{li2022languagedriven,yao2022detclip,radford2021learning} and fuses 2D prediction results into 3D if required. Though promising, they suffer from significant computation and storage overheads during inference and can be sub-optimal to address 3D understanding without learning from 3D geometries.

This paper focuses on another open-world research direction that concentrates on \textit{open-world point cloud learning}. It requires training 3D backbones to enable their open-world capabilities without the image modality dependence during inference and thus have more applicability potential.
Along this line of research, some~\cite{zeng2023clip2,peng2022openscene,zhang2023clip} have attempted to distill 2D dense features~\cite{li2022languagedriven,yao2022detclip} into 3D backbones for 3D feature learning. However, they still incur high training costs and might inherit 2D prediction failure modes.
In addition, Ding {\etal} ~\cite{ding2022language,ding2023lowis3d} obtain point-language paired data through image captioning by VL foundation models for training 3D backbones.
These methods are scalable toward a large vocabulary space and can be easily integrated with advanced 3D backbones.
Despite the advantages, they still suffer from the coarse text supervision. 

\begin{figure*}
    \vspace{-0.3cm}
    \centering
    \includegraphics[width=1\linewidth]{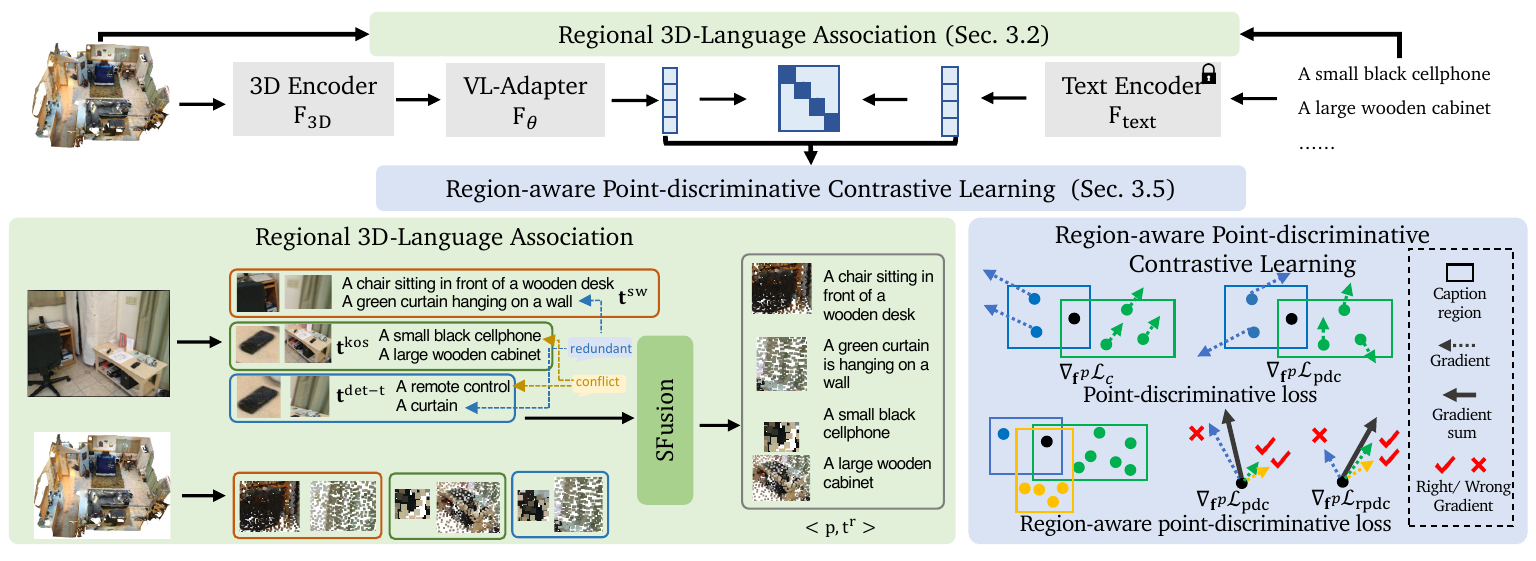}
    \vspace{-0.7cm}
    \caption{Overview of our regional point-language contrastive learning framework. For regional 3D-language association, We develop a 3D-aware SFusion strategy effectively combining 3D vision-language pairs obtained from multiple 2D foundation models (refer to Sec.~\ref{sec:region_prompted_language}). Upon these 3D-language data, we propose region-aware point-discriminative contrastive learning to facilitate more distinctive and robust representation learning (detailed in Sec.~\ref{sec:pdc_loss}). Different point \& box colors in the bottom-right indicate various 3D-caption pairs. }
    \label{fig:framework}
    \vspace{-10pt}
\end{figure*}

\section{RegionPLC}
\vspace{-0.1cm}
\subsection{Overview}
\vspace{-0.1cm}
We focus on 3D open-world scene understanding at both semantic and instance levels. 
During training, given a point cloud of a scene $\mathcal{P}=\{\textbf{p}\}$, the model can utilize human annotations $\mathcal{Y}$ for base categories $\mathcal{C}^B$, but cannot access annotations for novel categories $\mathcal{C}^N$.
During the inference phase, the trained model needs to classify and localize points associated with both base and novel categories ($\mathcal{C}^B \cup \mathcal{C}^N$).

To achieve open-world understanding, apart from the common 3D encoder $\text{F}_{\text{3D}}$, we follow \cite{ding2022language} to replace the classification layer weights with category embeddings $\mathbf{f}^l$ extracted from a pretrained text encoder $\text{F}_{\text{text}}$ of CLIP~\cite{radford2021learning} (See Figure~\ref{fig:framework}: Upper). 
Hence, the prediction process is shown as follows:
{\setlength\abovedisplayskip{3.0pt}
\setlength\belowdisplayskip{3.1pt}
\begin{equation}
    \mathbf{f}^p=\text{F}_{\theta}({\text{F}_{\text{3D}}(\mathbf{p})}), ~
    \mathbf{s}= \sigma(\mathbf{f}^l\cdot \mathbf{f}^p), ~
    \mathbf{o}=\text{F}_{\text{loc}}(\text{F}_{\text{3D}}(\mathbf{p}),  \mathbf{s}),
\end{equation}
}where $\text{F}_{\theta}$ is the vision-language (VL) adapter to align the feature dimension of the 3D point-wise features $\mathbf{f}^p$ and category embeddings $\mathbf{f}^l$, $\mathbf{s}$ is the semantic classification score, $\sigma$ is the softmax function, $\mathbf{o}$ is the instance proposal output, and $\text{F}_\text{loc}$ is the localization network~\cite{vu2022softgroup} for instance segmentation.
With these modifications, 
the model can predict any desired categories by computing similarity between point-wise features and queried category embeddings for open-world inference. 

The goal of our RegionPLC is to train such an open-world 3D backbone via dense region-level 3D-language supervision leveraging powerful and diverse 2D foundation models, as shown in Figure~\ref{fig:framework}. 
we first obtain region-level 2D-language pairs through three streams of 2D VL models (\ie image captioning~\cite{wang2022ofa}, object detection~\cite{zhou2022detecting} and dense captioning~\cite{wu2022grit,peng2023kosmos2}) and then associate them to 3D points (see Sec.~\ref{sec:region_prompted_language}). 
Then, we comprehensively benchmark and examine these 3D-language pairs from different sources, deriving their merits and shortcomings for 3D learning (see Sec.~\ref{sec:analysis}).  
Based on our study, we propose a simple \textbf{S}upplementary-oriented \textbf{Fusion} (\textbf{SFusion}) strategy leveraging their 3D relationships to alleviate redundancies and conflicts in Sec.~\ref{sec:caption_integration}, obtaining vocabulary-enriched and denser region-level 3D-language paired data.
Finally, upon the 3D-language data, we design a region-aware point-discriminative contrastive learning objective to replace CLIP-style loss for more robust and discriminative feature learning from language supervisions in Sec.~\ref{sec:pdc_loss}. 



\subsection{Regional 3D-Language Association from 2D Foundation Models }\label{sec:region_prompted_language}
\vspace{-0.1cm}
Here, we first introduce three streams of methods along with two types of visual prompts to extract regional language descriptions from 2D vision-language foundation models: $i$) object detector with language template; $ii$) explicit visual prompted image captioning; $iii$) dense captioning.

\begin{figure}
    \centering
    \vspace{-0.2cm}
    \includegraphics[width=0.95\linewidth]{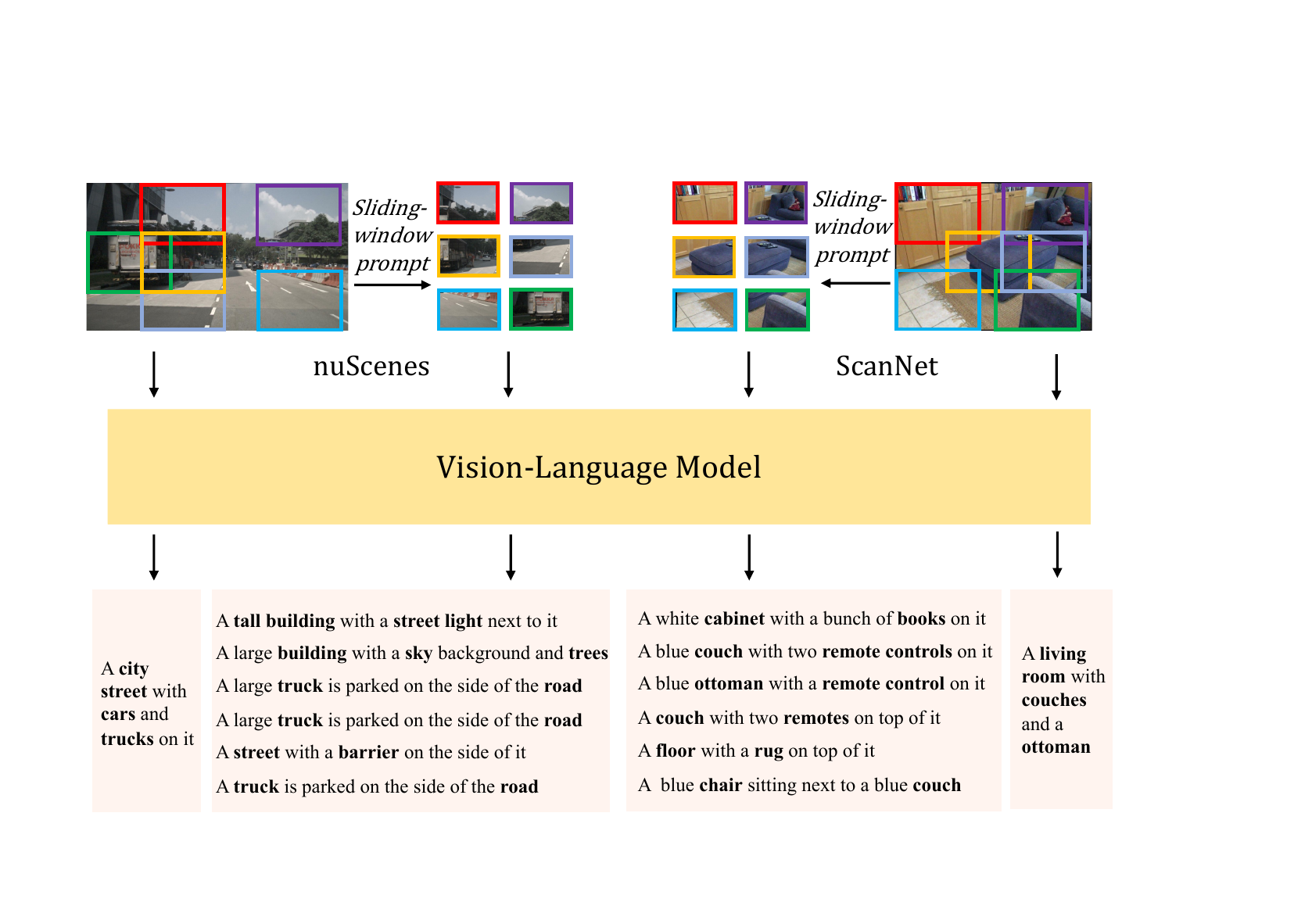}
    \vspace{-0.4cm}
    \caption{Comparisons of different advanced manners for extracting regional language descriptions with 2D foundation models.} 
    \vspace{-0.7cm}
    \label{fig:caption_compare}
\end{figure}

\vspace{0.05in}\noindent\textbf{Object Detector with language template $\mathcal{G}^\text{det}$.} The most straightforward manner to obtain regional language supervision is to leverage the category prediction from 2D object detector Detic~\cite{zhou2022detecting} and then fill the category into a language template as CLIP~\cite{radford2021learning}, as illustrated in Figure~\ref{fig:caption_compare}. Thanks to the multi-scale training strategy, object detectors can capture remote and small objects. We denote such regional captions as $\mathbf{t}^{\text{det-t}}$.

\vspace{0.05in}\noindent\textbf{Explicit visual prompted image captioning $\mathcal{G}^\text{prompt}$.}
Another intuitive paradigm is first to generate explicit visual prompts such as boxes and then caption these image patches via image captioning model OFA~\cite{wang2022ofa} (refer to Figure~\ref{fig:caption_compare}).
As for obtaining explicit visual prompts, we attempt two types: sliding windows and object proposals. Sliding-window-cropped image patches cover all potential semantic regions without being constrained by pre-defined vocabulary space, benefiting open-world tasks but sacrificing precise localization. In contrast, 2D object proposals from detectors provide more accurate object localization but suffer from the limited vocabulary space with pre-defined label space such as LVIS~\cite{gupta2019lvis}.
The obtained dense region-level captions through sliding-window prompts and detector prompts are denoted as $\mathbf{t}^{\text{sw}}$ and $\mathbf{t}^{\text{det-c}}$, respectively.

\begin{table}[htbp]
    \vspace{-0.1cm}
    \centering
    \begin{small}
    \setlength\tabcolsep{2pt}
    \scalebox{0.87}{
        \begin{tabular}{l|c|c|c|c}
            \bottomrule[1pt]
            \multirow{2}{*}{Method}  & \multicolumn{2}{c|}{ScanNet B12/N7} & \multicolumn{2}{c}{ScanNet annotation-free} \\
            \cline{2-5}
            & 25K & 125K &  25K & 125K \\
            \hline
            $\mathbf{t}^\text{det-t}$ & 63.4 / \textbf{70.3} / 57.7 & \textbf{67.4} / 70.5 / \textbf{64.6} & 37.7 (59.7) & 41.9 (64.5) \\
            $\mathbf{t}^\text{sw}$ & \textbf{65.9} / 70.2 / \textbf{62.1} & 66.0 / 70.2 / 62.3 & 48.1 (69.2) & 47.8 (69.2) \\
            $\mathbf{t}^\text{det-c}$ & 64.4 / 69.9 / 59.7 & 65.6 / 70.7 / 61.2 & 41.4 (64.1) & 43.8 (65.6) \\
            $\mathbf{t}^\text{grit}$ & 62.7 / \textbf{70.3} / 56.6 & 64.9 / \textbf{70.8} / 59.9 & \textbf{50.5} (\textbf{72.2}) & 51.3 (\textbf{74.2}) \\
            $\mathbf{t}^\text{kos}$ & 64.4 / \textbf{70.3} / 59.4 & 64.6 / 69.8 / 60.2 & 50.1 (70.7) & \textbf{51.7} (72.7) \\
            \hline
            $\mathbf{t}^\text{kos} \cup \mathbf{t}^\text{det-t}$ & 64.6 / 69.8 / 60.2 & \textbf{67.0} / 69.9 / \textbf{64.4} & 44.4 (66.3) & \textbf{54.0} (\textbf{74.5}) \\
            $\mathbf{t}^\text{kos} \cup \mathbf{t}^\text{sw}$ & \textbf{65.9} / 69.9 / \textbf{62.4} & 65.6 / \textbf{70.9} / 61.0 & \textbf{53.5} (72.9) & 53.1 (73.9) \\
             $\mathcal{L}_{\mathbf{t}^\text{kos}}$ + $\mathcal{L}_{\mathbf{t}^\text{det-t}}$ & 65.0 / \textbf{70.2} / 60.5 & 64.4 / 69.1 / 60.2 & 51.3 (70.7) & 51.2 (72.4) \\
            $\mathcal{L}_{\mathbf{t}^\text{kos}}$ + $\mathcal{L}_{\mathbf{t}^\text{sw}}$ & 65.4 / 70.0 / 61.4 & 64.6 / 70.2 / 59.8 & 52.9 (\textbf{73.6}) & 52.7 (73.8) \\
            \toprule[0.8pt]
        \end{tabular}
     }
    \end{small}
    \vspace{-0.3cm}
    \caption{Results of regional caption fusion on base-annotated (hIoU / mIoU$^\mathcal{B}$ / mIoU$^\mathcal{N}$) and annotation-free (mIoU$^\dagger$ (mAcc$^\dagger$), tested on foreground classes only) 3D ScanNet semantic segmentation. $\mathbf{t}^\text{kos} \cup \mathbf{t}^\text{det-t}$ and $\mathcal{L}_{\mathbf{t}^\text{kos}}$ + $\mathcal{L}_{\mathbf{t}^\text{det-t}}$ indicate data-level and multi-loss fusion, respectively. Best results are presented in \textbf{bold}.}
    \label{tab:benchmark}
    \vspace{-0.6cm}
\end{table}

\noindent\textbf{Dense Captioning $\mathcal{G}^\text{cap}$.}
Apart from the powerful image detectors and image captioning models, recent advances in dense captions and grounding models such as GRiT~\cite{wu2022grit} and Kosmos-2~\cite{peng2023kosmos2} are trained on the large-scale 2D box and box description pairs. As shown in Figure~\ref{fig:caption_compare}, dense captioners offer precise object localization and rich vocabulary spaces but tend to focus on only salient objects and ignore small and distant objects.
We denote captions generated through GRiT~\cite{wu2022grit}, Kosmos-2~\cite{peng2023kosmos2} and Detic~\cite{zhou2022detecting} with a caption template as $\mathbf{t}^\text{grit}$, and $\mathbf{t}^\text{kos}$ and $\mathbf{t}^\text{det-t}$, respectively.

\vspace{0.02in}
\noindent\textbf{Associate Points to Dense Captions.}~\label{sec:associate}
Upon above 5 types of regional captions $\mathbf{t}^r = \{\mathbf{t}^\text{sw}, \mathbf{t}^\text{det-c}, \mathbf{t}^\text{det-t}, \mathbf{t}^\text{grit}, \mathbf{t}^\text{kos}\}$, we associate them to partial point sets through 3D geometry, similar to~ \cite{ding2022language,peng2022openscene}, to pair points and language. Specifically, we begin by projecting the 3D scenes onto 2D images to align points with pixels. Then by connecting the points $\mathbf{\hat{p}}$ within each 2D region to their respective captions, we obtain the regional 3D-language pairs $\langle\mathbf{\hat{p}}, \mathbf{t}^r\rangle$.

\vspace{-0.2cm}
\subsection{Benchmark and Analysis on Regional 3D-Language Pairs}\label{sec:analysis}
\vspace{-0.1cm}

With the constructed five types of regional 3D-language pairs $\mathbf{t}^r = \{\mathbf{t}^\text{sw}, \mathbf{t}^\text{det-c}, \mathbf{t}^\text{det-t}, \mathbf{t}^\text{grit}, \mathbf{t}^\text{kos}\}$, the follow-up question is which delivers the best performance on learning 3D open-world representation and how to combine them to obtain enriched vocabulary space and denser regional 3D-language association. 
Hence, we benchmark them on ScanNet~\cite{dai2017scannet} semantic segmentation tasks with different novel categories and 2D image quantities (25K \vs 125K).
Our benchmark encompasses two settings: $i$) the B12/N7 setting including 12 annotated base categories and 7 unannotated novel categories, which requires a strong comprehension of a large vocabulary corpus; $ii$) the annotation-free setting, wherein all categories are novel ones, and thus necessitates both open-vocabulary recognition and precise object localization with only sparse 3D-language pairs.

\noindent\textit{\textbf{Complementary cues.}} As shown in the upper of Table~\ref{tab:benchmark}, no single type of 3D-language source consistently outperforms others in all settings, and each association has its own merits. For example, $\mathbf{t}^\text{det-t}$ inherits the advanced small object localization capabilities (refer to Figure~\ref{fig:caption_compare} middle for ``traffic light'' and ``wheel'' descriptions.), excelling others in the ScanNet B12/N7 (125K). However, it suffers from the limited pre-defined vocabulary space and obtains the worst performance in the annotation-free setting with 17 novel categories. 
In contrast, dense captioners $\mathbf{t}^\text{kos}$ and $\mathbf{t}^\text{grit}$ offer salient object localization with semantic-rich vocabulary (refer to Figure~\ref{fig:caption_compare} right for attribute descriptions), exhibiting superior results on the annotation-free setting. This suggests that different VL models and visual prompts offer various merits and might complement each other. 



\noindent\textit{\textbf{End-to-end manners scale better}}. When comparing the performance of utilizing 25K and 125K images, we find that end-to-end trained dense captioners and detectors (\ie $\mathbf{t}^\text{kos}$, $\mathbf{t}^\text{grit}$ and $\mathbf{t}^\text{det-t}$) scale better than the two-stage image captioning manners with visual prompts. The reason might be that end-to-end trained dense caption sources are more consistent on different views and thus yield fewer semantic conflicts when scaling up to more views. 

\noindent\textit{\textbf{Common combinations are not always effective.}} 
As above-mentioned, different 3D-language pairs can offer complementary cues. Hence, we examine their synergy effect for better performance. As shown in the bottom of Table~\ref{tab:benchmark}, we attempt to combine the representatives from three streams of regional caption generation manners $\mathbf{t}^\text{kos}, \mathbf{t}^\text{det-t}$ and $\mathbf{t}^\text{sw}$ via data-level and multi-loss fusion. 
Nevertheless, the performance lift across different settings is not consistent or only shows incremental increases, which suggests the need for a more dedicated fusion strategy to accommodate extensive dense language supervision from multiple sources.


\vspace{-0.1cm}
\subsection{Boost Synergy of Diverse 3D-language Sources}\label{sec:caption_integration}
\vspace{-0.1cm}
Motivated by the observations of complementary merits of individual 3D-language sources and their unsatisfactory synergy results, we further study how to combine these varied 3D-language sources effectively and efficiently.
In this regard, we propose a Supplementary-orientated Fusion (SFusion) strategy to integrate the most diverse semantic clues while filtering out potential conflicts from different caption sources. 
As data-level mixing delivers better performance than loss-level combination, we focus on tackling the bottleneck of data-level 3D-language pairs fusion here.
When training 3D models on data-level mixed 3D-language pairs, they are learning from a more informative language description, but suffer from sub-optimal performance. This suggests that the main challenges in straightforward data-level mixing are the redundancy and conflicts from different caption sources, especially for highly overlapped point cloud regions (see Figure~\ref{fig:framework}).
For those highly overlapped 3D regions with multiple language sources, mutually conflicting descriptions will confuse models, and the overabundance of repetitive language descriptions tends to overwhelm optimization toward easily identifiable areas, leading to sub-optimal performance.

Hence, our SFusion addresses these problems by fusing 3D-language pairs with low 3D overlaps to alleviate potential conflicts in overlapped areas and obtain spatially supplementary 3D-language pairs.
Specifically, we first select the most reliable caption source that performs best as the primary 3D-language source $\mathbf{t}^\text{pri}$. Then, we compute the overlap ratio $\tau$ of point sets between the primary source and candidate caption sources $\mathbf{t}^\text{can}$ on $i$-th 3D scene as follows,
\vspace{-0.3cm}
\begin{equation}\vspace{-0.3cm}
    \tau_{jk} = \text{overlap}(\mathbf{\hat{p}}_{ij}^\text{pri}, \mathbf{\hat{p}}_{ik}^\text{can}), ~~~~~ \hat{\tau}_{k} = \max_j \tau_{jk},
\end{equation}
where $\text{overlap}$ measures the intersection over union (IoU) between two point sets, $\mathbf{\hat{p}}_{ij}^\text{pri}$ and $\mathbf{\hat{p}}_{ik}^\text{can}$ are the $j$-th and $k$-th point set in the $i$-th scene from the primary source and candidate source, respectively.
Then, we define thresholds $T_{l}$ and $T_{h}$ to filter out 3D-language pairs with high overlap ratios from $\mathbf{t}^\text{can}$, which might result in redundant or conflict supervision to the primary source. 
Hence, only candidate 3D-language pairs $\langle\mathbf{\hat{p}}_{ik}^\text{can}, \mathbf{t}_{ik}^\text{can}\rangle$ with $ T_l <\hat{\tau}_{k} < T_h$ ($T_{l}$ set to zero) are fused with the primary source.
This procedure can be iteratively applied across all candidate caption sources to obtain a collection of 3D-language pairs with low geometrical overlaps. 
This refined set will serve as the supervision for the follow-up contrastive training.
Notice that we also introduce a hyper-parameter $\epsilon \in [0, 1]$ to control the ratio of the primary source and candidate source during fusion, as maintaining the majority of primary sources is beneficial during training with multi-source 3D-language pairs.
Our experimental results in Table~\ref{tab:sfusion} verify our above claims and demonstrate that our SFusion strategy can significantly boost the combination 
of multiple language sources.

\subsection{Region-aware Point-discriminative Contrastive Learning}\label{sec:pdc_loss}
After obtaining 3D-language pairs $\langle\mathbf{\hat{p}}, \mathbf{t}\rangle$ for supervision, we proceed to train $\mathbf{F}_\text{3D}$ and $\mathbf{F}_\theta$ to align 3D features with language features for open-world learning. We introduce region-aware point-discriminative contrastive loss as below. 

\vspace{0.05in}
\noindent\textbf{CLIP-style Contrastive Loss.}
\footnotetext[1]{\url{https://kaldir.vc.in.tum.de/scannet_benchmark/documentation}}
We can pull paired 3D features and language features closer while pushing away the unmatched ones through 
CLIP-style~\cite{radford2021learning} contrastive loss (refer to Figure~\ref{fig:framework} top right). It can be formulated as:
{\setlength\abovedisplayskip{3.0pt}
\setlength\belowdisplayskip{3.1pt}
\begin{gather}\label{eq:clip_contrast}
    \mathbf{f}^{\hat{p}} = \text{Pool}(\mathbf{\hat{p}}, \mathbf{f}^p), ~~\mathbf{\hat{z}}=\mathbf{f}^{\hat{p}}\cdot\mathbf{F}^t, ~~ \mathbf{\hat{s}}=\sigma(\mathbf{\hat{z}}), \\
    \mathcal{L}_{c}=- \mathbf{y}^t\cdot\ln \mathbf{\hat{s}}
    ,
\end{gather}}where $\mathbf{f}^{\hat{p}}$ is the average-pooled region feature, Pool($\mathbf{\hat{p}},\mathbf{f}^{\hat{p}}$) is our custom CUDA operator to gather features $\mathbf{f}^{\hat{p}}$ over point set $\mathbf{\hat{p}}$, $\mathbf{F}^t=[\mathbf{f}^t_1, \mathbf{f}^t_2,\cdots\mathbf{f}^t_{n_t}]$ concatenates all caption embeddings in a scene, $\mathbf{\hat{z}}$ and $\mathbf{\hat{s}}$ measure the similarity and the score probability between a 3D region and all captions, $\sigma$ is sigmoid function and $\mathbf{y}^t$ is the one-hot label highlighting the position paired with $\mathbf{\hat{p}}$. 
While CLIP~\cite{radford2021learning} targets learning a global image-level feature for the classification task, it neglects the demand of learning point-wise discriminative features for dense prediction tasks. As shown in Figure~\ref{fig:framework}, the pooling operation will average the point-wise features and make all points in the same region $\mathbf{\hat{p}}$ optimized in the same direction, preventing the learning of discriminative representations for dense prediction tasks.
We present more analysis on this undesired effect in the suppl..

\vspace{0.05in}\noindent\textbf{Point-discriminative Contrastive Loss.}
Considering the limitation of CLIP-style loss, we propose a point-discriminative contrastive loss $\mathcal{L}_\text{pdc}$  to make the learning of point embedding discriminative. 
Specifically, for each regional 3D-language pair, instead of aggregating point features into an averaged region-level feature, our $\mathcal{L}_\text{pdc}$ directly computes the similarity between point-wise embeddings and caption embeddings. We then pool the logarithm of predicted point-wise probability within $\mathbf{\hat{p}}$ to compute the cross-entropy loss regarding one-hot label $\mathbf{y}^t$ as follows,
{\setlength\abovedisplayskip{3.0pt}
\setlength\belowdisplayskip{3.1pt}
\begin{gather}
    \mathbf{z}=\mathbf{f}^p\cdot\mathbf{F}^t, ~~ \mathbf{s}=\sigma(\mathbf{z}), ~~ \mathcal{L}_{\text{pdc}}=-\mathbf{y}^t \cdot \text{Pool}(\mathbf{\hat{p}}, \ln \mathbf{s}), 
\end{gather}}where $\mathbf{z}$ and $\mathbf{s}$ indicate the similarity and probability matrix between point-wise features and all caption embeddings. By doing so, the optimization direction of each point will be adapted to its own point embeddings and thus make them discriminative (refer to Figure \ref{fig:framework}).  More details are included in the supplementary materials.

\vspace{0.05in}\noindent\textbf{Region-aware Normalization.}
Though discriminative, the $\mathcal{L}_\text{pdc}$ will back-propagate smaller gradients to points in large regions due to the pooling operation, leading to an implicit bias towards region size which can be harmful to representation learning. 
To alleviate this issue, we propose a region-aware factor to normalize $\mathcal{L}_\text{pdc}$ by the region size, to ensure an equivalent gradient scale on points in each region regardless of its size.Obtained region-aware loss $\mathcal{L}_\text{rpdc}$ as follows,
\vspace{-0.15cm}
\begin{equation}\vspace{-0.15cm}
    \mathcal{L}_{\text{rpdc}}=- \alpha_r \mathcal{L}_{\text{pdc}}, ~~ \alpha_r = \frac{n_t\cdot\text{card}(\mathbf{\hat{p}})}{\sum^{n_t}_i\text{card}(\mathbf{\hat{p}}_i)},
\end{equation}
where $\alpha_r$ is the region-aware normalization factor, $n_t$ is the number of 3D-language pairs each scene.

\begin{table*}[htbp]
    \vspace{-0.4cm}
    \centering
    \begin{small}
    \setlength\tabcolsep{3.5pt}
    \scalebox{0.92}{
        \begin{tabular}{l|c|c|c|c|c|c|c}
            \bottomrule[1pt]
            \multirow{2}{*}{Method}  & \multicolumn{3}{c|}{ScanNet~\cite{dai2017scannet}} & \multicolumn{2}{c|}{nuScenes~\cite{caesar2020nuscenes}} & \multicolumn{2}{c}{ScanNet200~\cite{rozenberszki2022language}}\\
            \cline{2-8}
            & {B15/N4} & {B12/N7} & {B10/N9} & {B12/N3} & {B10/N5} & {B170/N30} & {B150/N50} \\
            \hline
            3DGenZ~\cite{michele2021generative} & 20.6 / 56.0 / 12.6 & 19.8 / 35.5 / 13.3 & 12.0 / 63.6 / 6.6 & 1.6 / 53.3 / 0.8 & 1.9 / 44.6 / 1.0 & 2.6 / 15.8 / 1.4 & 3.3 / 14.1 / 1.9 \\
            3DTZSL~\cite{cheraghian2020transductive} & 10.5 / 36.7 / 6.1 & 3.8 / 36.6 / 2.0 & 7.8 / 55.5 / 4.2 & 1.2 / 21.0 / 0.6 & 6.4 / 17.1 / 3.9 & 0.9 / 4.0 / 0.5 & 0.7 / 3.8 / 0.4 \\
            OVSeg-3D~\cite{ding2022language} & 0.0 / 64.4 / 0.0 & 0.9 / 55.7 / 0.1 & 1.8 / 68.4 / 0.9 & 0.6 / 74.4 / 0.3 & 0.0 / 71.5 / 0.0 & 1.5 / 21.1 / 0.8 & 3.0 / 20.6 / 1.6\\
            PLA~\cite{ding2022language} & 65.3 / \textbf{68.3} / 62.4 & 55.3 / 69.5 / 45.9 & 53.1 / 76.2 / 40.8 & 47.7 / 73.4 / 35.4 & 24.3 / 73.1 / 14.5 & 11.4 / 20.9 / 7.8 & 10.1 / 20.9 / 6.6 \\
            \hline
            RegionPLC & \textbf{69.4} / 68.2 / \textbf{70.7} & \textbf{68.2} / \textbf{69.9} / \textbf{66.6} & \textbf{64.3} / \textbf{76.3} / \textbf{55.6} & \textbf{64.4} / \textbf{75.8} / \textbf{56.0} & \textbf{49.0} / \textbf{75.8} / \textbf{36.3} & \textbf{16.6} / \textbf{21.6} / \textbf{13.9} & \textbf{14.6} / \textbf{22.4} / \textbf{10.8} \\
            \hline
            Fully-Sup. & 73.3 / 68.4 / 79.1 & 70.6 / 70.0 / 71.8 & 69.9 / 75.8 / 64.9 & 73.7 / 76.6 / 71.1 & 74.8 / 76.8 / 72.8 & 20.9 / 21.7 / 20.1 & 20.6 / 22.0 / 19.4 \\
            \toprule[0.8pt]
        \end{tabular}
     }
    \end{small}
    \vspace{-0.4cm}
    \caption{Results for open-world 3D semantic segmentation on ScanNet, nuScenes and ScanNet200 in terms of hIoU / mIoU$^\mathcal{B}$ / mIoU$^\mathcal{N}$. Best open-world results are presented in \textbf{bold}.}
    \label{tab:sem_seg}
    \vspace{-0.6cm}
\end{table*}

\noindent\textbf{Analysis.} With point-discriminative and region-aware properties, our $\mathcal{L}_{\text{rpdc}}$ facilitates more superior and robust representation learning. It allows each point to grasp its unique semantics without disruptions from other unrelated points (refer to Figure~\ref{fig:framework} right). This is especially vital for annotation-free dense prediction to segment object boundaries without any annotation (see Table~\ref{tab:component} for verification). 
Moreover, the region-aware factor in $\mathcal{L}_{\text{rpdc}}$ provides a more robust optimization procedure. 
As depicted in the right section of Figure~\ref{fig:framework}, points associated with multiple captions are normalized to a similar gradient scale. When multiple captions reach a consensus, this leads to consistent gradient directions, thereby encouraging them to be optimized in a unified direction. 
Conversely, when multiple captions conflict, this leads to inconsistent gradient directions and thus discourages the noisy optimization.

\vspace{-0.2cm}
\section{Experiments}
\vspace{-0.1cm}
\subsection{Basic Setups}
\vspace{-0.1cm}
\noindent\textbf{Datasets and Validation Settings.} To test the effectiveness of RegionPLC, we evaluate it on three popular datasets: ScanNet~\cite{dai2017scannet}, ScanNet200~\cite{rozenberszki2022language} and nuScenes~\cite{caesar2020nuscenes}, covering indoor and outdoor scenarios.
We validate the open-world capability of our method with different numbers of annotated categories, including \textbf{base-annotated open world} (\ie part of categories annotated) and \textbf{annotation-free open world} (\ie no category annotated). We evaluate our method's performance on both semantic segmentation and instance segmentation tasks.

\vspace{0.05in}\noindent\textbf{Category Partition.}
We split categories into base and novel on ScanNet~\cite{dai2017scannet} following PLA~\cite{ding2022language}. For nuScenes~\cite{caesar2020nuscenes}, we ignore the ``otherflat" class and randomly divide the rest classes into B12/N3 (\ie 12 base and 3 novel categories) and B10/N5. For ScanNet200~\cite{rozenberszki2022language}, we randomly split 200 classes to B170/N30 and B150/N50. See Suppl. for details.

\vspace{0.05in}\noindent\textbf{Evaluation Metrics.} For semantic segmentation, we follow \cite{xian2019semantic,ding2022language} to employ mIoU$^\mathcal{B}$, mIoU$^\mathcal{N}$ and harmonic mean IoU (hIoU) for evaluating base, novel categories and their harmonic mean separately. Similarly, for instance segmentation, we employ mAP$_{50}^\mathcal{B}$, mAP$_{50}^\mathcal{N}$ and hAP$_{50}$.For annotation-free semantic segmentation, we use mean IoU and mean accuracy on foreground classes (\ie mIoU$^\dag$ and mAcc$^\dag$) excluding ``wall'', ``floor'' and ``ceiling'' for evaluation.

\vspace{0.05in}\noindent\textbf{Implementation Details.} We adopt the sparse-convolution-based UNet~\cite{graham20183d} as the 3D encoder with CLIP~\cite{radford2021learning} text encoder as the final classifier for 3D semantic segmentation, and SoftGroup~\cite{vu2022softgroup} for instance segmentation as \cite{ding2022language}. We use category prompts to replace ambiguous category names such as ``manmade'' and ``drivable surface'' with a list of concrete category names when encoding category embeddings.
We run all experiments with a batch size of 32  on 8 NVIDIA V100 or A100 (see Suppl. for more details).

\vspace{-0.1cm}
\subsection{Base-annotated Open World}
\vspace{-0.1cm}
\noindent\textbf{Comparison Methods.} We compare RegionPLC to previous open-world or zero-shot works.
3DGenZ~\cite{michele2021generative} and 3DTZSL~\cite{cheraghian2020transductive} are early works for 3D zero-shot learning reproduced by \cite{ding2022language}. OVSeg-3D extends LSeg to 3D~\cite{li2022languagedriven}, reported by \cite{ding2022language}. PLA~\cite{ding2022language} is the previous cutting-edge method.

\vspace{0.05in}\noindent\textbf{3D Semantic Segmentation.} As shown in Table~\ref{tab:sem_seg}, compared to the previous state-of-the-art method PLA~\cite{ding2022language}, our method largely lifts the mIoU of unseen categories by $8.3\% \sim 21.8\%$ among various partitions on ScanNet and nuScenes. Furthermore, when compared to baselines without language supervision, \ie 3DGenZ~\cite{michele2021generative} and 3DTZSL~\cite{cheraghian2020transductive}, our method even obtains $30.6\% \sim 42.7\%$ performance gains regarding mIoU on novel categories among different partitions and datasets. These significant and consistent improvements across indoor and outdoor scenarios show the effectiveness of our RegionPLC framework.

Furthermore, when facing more long-tail dataset ScanNet200~\cite{rozenberszki2022language}, our method still obtains notable mIoU$^\mathcal{N}$ gains ranging from $4.2\%$ to $5.1\%$ compared to PLA~\cite{ding2022language} as shown in Table~\ref{tab:sem_seg}. In this regard, our proposed region-level language supervision and region-aware point-discriminative contrastive loss show its potential to address 3D open-world understanding in complex and long-tail scenarios.

\begin{table}[h]
    \vspace{-0.3cm}
    \centering
    \begin{small}
    \setlength\tabcolsep{2pt}
    \scalebox{0.9}{
        \begin{tabular}{c|c|c|c}
            \bottomrule[1pt]
            {Method} & \multicolumn{3}{c}{ScanNet}  \\
            \cline{2-4}
            & B13/N4 & B10/N7 & B8/N9 \\
            \hline
            OVSeg-3D~\cite{li2022languagedriven}& 5.1 / 57.9 / 2.6 & 2.0 / 50.7 / 1.0 & 2.4 / 59.4 / 1.2  \\ 
            PLA~\cite{ding2022language} & 55.5 / 58.5 / 52.9 & 31.2 / \textbf{54.6} / 21.9 & 35.9 / \textbf{63.1} / 25.1 \\
            \hline
            RegionPLC  & \textbf{58.2} / \textbf{59.2} / \textbf{57.2} & \textbf{40.6} / 53.9 / \textbf{32.5} & \textbf{46.8} / 62.5 / \textbf{37.4} \\
            \hline
            Fully-Sup. & 64.5 / 59.4 / 70.5 & 62.5 / 57.6 / 62.0 & 62.0 / 65.1 / 62.0 \\
            \toprule[0.8pt]
        \end{tabular}
     }
    \end{small}
    \vspace{-0.4cm}
    \caption{Results for open-world 3D instance segmentation on ScanNet in terms of hAP$_{50}$ / mAP$_{50}^\mathcal{B}$ / mAP$_{50}^\mathcal{N}$.}
    \vspace{-0.4cm}
    \label{tab:inst_seg}
\end{table}

\begin{table*}[htbp]
    \vspace{-0.5cm}
    \centering
    \begin{small}
    \setlength\tabcolsep{5pt}
    \scalebox{0.85}{
        \begin{tabular}{l|l|cc|ccccc}
            \bottomrule[1pt]
            Method & Network & mIoU$^{\dag}$ & mAcc$^{\dag}$ & \makecell{Multi-view Infer} & \makecell{GT Instance Mask} & Train Hours & Extra Storage & Latency \\
            \hline
            MaskCLIP$^\ddag$~\cite{zhou2022maskclip} & CLIP~\cite{radford2021learning} & 23.1 & 40.9 & $\checkmark$ & $\times$ & - & - & 1.7 s \\
            OpenScene-2D~\cite{peng2022openscene} & LSeg~\cite{li2022languagedriven} & 58.0 & 68.5 & $\checkmark$ & $\times$ & - & - & 106.1s \\
            \hline
            OpenScene-3D$^\ddag$~\cite{peng2022openscene} & SparseUNet16~\cite{graham20183d} & 57.2 & 69.9 & $\times$ & $\times$ & 24.7 h & 117.3 G & \textbf{0.08 s} \\  
            OpenScene-3D$^\ddag$~\cite{peng2022openscene} & SparseUNet32~\cite{graham20183d} & 57.8 & 70.3 & $\times$ & $\times$ & 25.3 h & 117.3 G & 0.10 s \\
            PLA$^\ddag$~\cite{ding2022language} & SparseUNet16~\cite{graham20183d} & 17.7 & 33.5 & $\times$ & $\times$ & \textbf{11.5 h} & \textbf{1.1 G} & \textbf{0.08 s} \\
            PLA$^\ddag$~\cite{ding2022language} & SparseUNet32~\cite{graham20183d} & 19.1 & 41.5 & $\times$ & $\times$ & 12.0 h & \textbf{1.1 G} & 0.10 s \\
            RegionPLC & SparseUNet16~\cite{graham20183d} & 56.9 & 75.6 & $\times$ & $\times$ & 12.5 h & 5.5 G & \textbf{0.08 s} \\
            RegionPLC & SparseUNet32~\cite{graham20183d} & \textbf{59.6} & \textbf{77.5} & $\times$ & $\times$ & 13.0 h & 5.5 G & 0.10 s \\
            \hline
            RegionPLC + OpenScene-3D$^\ddag$ & SparseUNet16~\cite{graham20183d} & 60.1 & 74.4 & $\times$ & $\times$ & 25.9 h & 122.8 G & \textbf{0.08 s} \\
            RegionPLC + OpenScene-3D$^\ddag$ & SparseUNet32~\cite{graham20183d} & \textbf{63.6} & \textbf{80.3} & $\times$ & $\times$ & 26.4 h & 122.8 G & 0.10 s \\
            \hline
            Fully-Sup. & SparseUNet16~\cite{graham20183d} & 75.9 & 84.8 & $\times$ & $\times$ & 9.6 h & - & 0.08 s \\
            Fully-Sup. & SparseUNet32~\cite{graham20183d} & \textbf{77.9} & \textbf{86.2} & $\times$ & $\times$ & 10.5 h & - & 0.10 s \\
            \toprule[0.8pt]
        \end{tabular}
     }
    \end{small}
    \vspace{-0.4cm}
    \caption{Annotation-free 3D semantic segmentation on ScanNet. $^\ddag$ and $^\sharp$ mean results reproduced by us and Uni3D, independently.}
    \label{tab:zero-shot}
    \vspace{-0.5cm}
\end{table*}

\noindent\textbf{3D Instance Segmentation.} As our pipeline provides local language descriptions to fine-grained point sets and encourage points to learn discriminative features, it also benefits instance-level localization task. As shown in Table~\ref{tab:inst_seg}, our method consistently brings $4.3\% \sim 12.3\%$ gains compared to the state-of-the-art PLA~\cite{ding2022language} across three partitions on ScanNet. It is noteworthy that our method obtains more obvious improvements for partitions with fewer base categories (\ie B10/N7 and B8/N9), demonstrating the effectiveness of our RegionPLC in enabling the model to distinguish unseen instances without human annotations.    

\vspace{-0.1cm}
\subsection{Annotation-free Open World}
\vspace{-0.1cm}

\vspace{0.05in}\noindent\textbf{Comparison Methods.} As shown in Table~\ref{tab:zero-shot}, we compare two streams of methods: $i$) Training-free methods using multi-view images for inference~\cite{zhou2022maskclip,peng2022openscene}.
$ii$) Methods leveraging 2D vision-language models during training~\cite{peng2022openscene,ding2022language}.


\vspace{0.1cm}
\noindent\textbf{3D Semantic Segmentation.}
As shown in Table~\ref{tab:zero-shot}, our RegionPLC with SparseUNet32~\cite{graham20183d} backbone significantly outperforms all other competitive methods by $1.8\%\sim57.5\%$ mIoU$^\dag$ and $7.2\%\sim72\%$ mAcc$^\dag$. This is the first time that a 3D open-world model achieves state-of-the-art performance without any 3D annotation or 2D 
pixel-aligned image features but only sparse language supervision for learning.
Moreover, our RegionPLC can scale up by scaling the 3D backbone from SparseUNet16 to SparseUNet32, obtaining 2.6\% mIoU$^\dag$ gains, which shows the advantage of learning from sparse language supervision instead of pixel-aligned feature distillation from 2D encoders~\cite{peng2022openscene}. 
It is also noteworthy that RegionPLC can function as a lightweight plug-and-play module and thus be integrated with other methods such as OpenScene~\cite{peng2022openscene} to further boost about 4\% mIoU$^\dag$. Notably, our method is training-efficient, requiring less disk storage and training time compared to OpenScene.

\begin{table}[htbp]
    \centering
    \vspace{-0.3cm}
    \begin{small}
    \setlength\tabcolsep{3pt}
    \scalebox{0.9}{
        \begin{tabular}{ccc|c|c}
            \bottomrule[1pt]
            \cite{peng2022openscene} & \cite{ding2022language} & RegionPLC & RegionPLC + \cite{peng2022openscene} & Fully-Sup. \\
            \hline
            5.9 (10.2) & 1.8 (3.1) & \textbf{9.1} (\textbf{17.3}) & \textbf{9.6} (\textbf{17.8}) & 23.9 (32.9) \\
            \toprule[0.8pt]
        \end{tabular}
     }
    \end{small}
    \vspace{-0.3cm}
    \caption{Annotation-free open-world semantic segmentation on ScanNet200~\cite{rozenberszki2022language} in terms of mIoU$^\dag$ (mAcc$^\dag$).}
    \vspace{-0.5cm}
    \label{tab:zero-shot_scannet200}
\end{table}

\vspace{0.05in}\noindent\textbf{Long-tail Scenario.} 
As shown in Table~\ref{tab:zero-shot_scannet200}, we set up comparisons on the more challenging long-tail dataset ScanNet200~\cite{rozenberszki2022language}. Notably, our RegionPLC surpasses other counterparts by $3.2\%\sim7.4\%$ mIoU$^\dag$ and $7.1\%\sim14.2\%$ mAcc$^\dag$. 
Specifically, OpenScene is less effective on ScanNet200 with a large number of fine-grained categories as it inherits the shortcomings or bias of the 2D segmentation model that forgets a large number of concepts during fine-tuning, as verified in ~\cite{ding2022don,jatavallabhula2023conceptfusion}. 
In contrast, our RegionPLC directly learns in a rich vocabulary space with dense and diverse captions which
is closer to real open-world scenarios. 

\begin{figure*}[hbtp]
    \centering
    \vspace{-0.4cm}
    \includegraphics[width=1\linewidth]{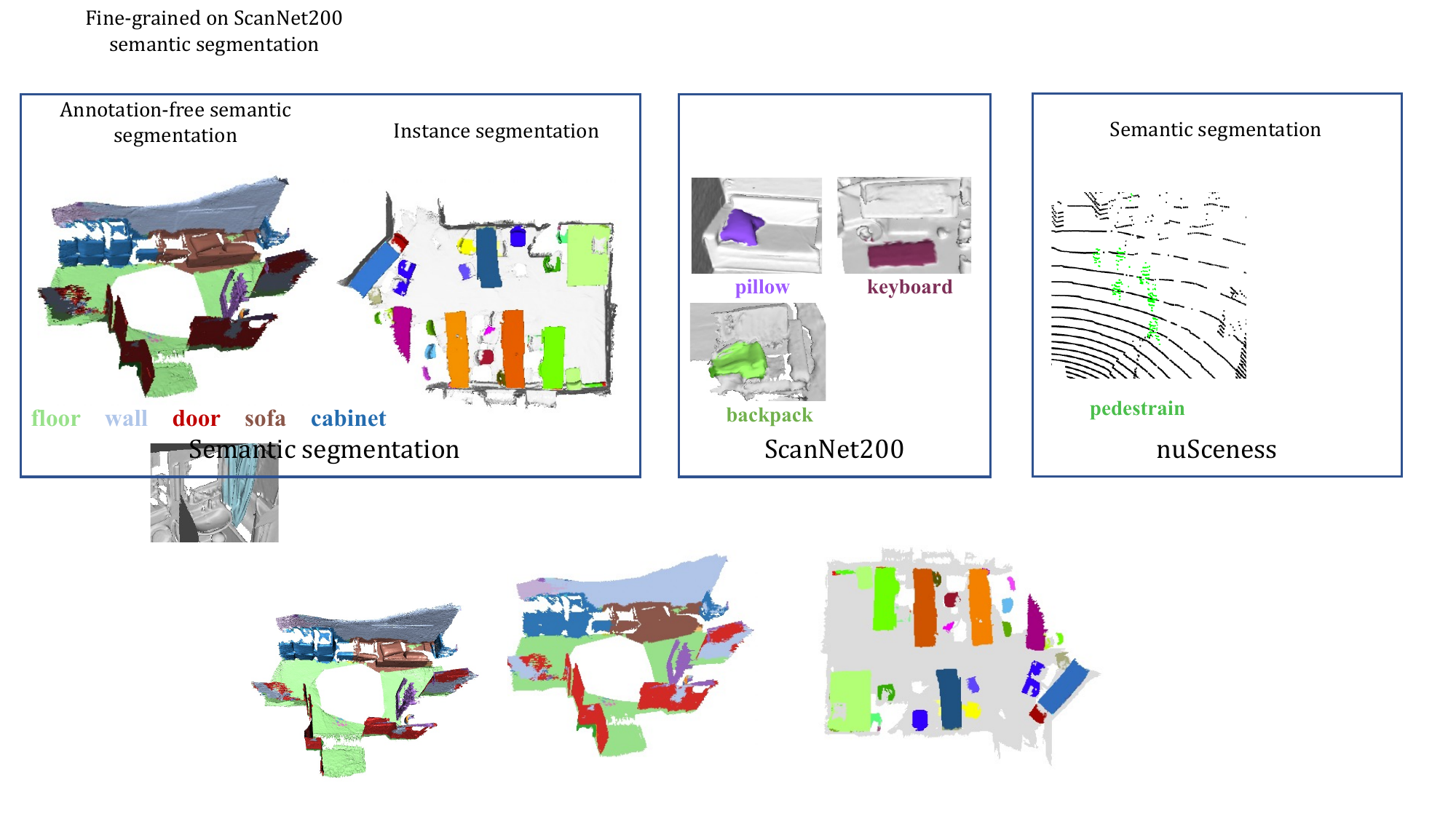}
    \vspace{-0.7cm}
    \caption{Qualitative results of our RegionPLC. The examples above show annotation-free open-world scene parsing where no human annotation is available (see (a)), and base-annotated open-world learning where a limited number of base classes are annotated (see (b), (c), (d)) for semantic and instance segmentation covering both indoor and outdoor scenarios. Unseen categories are highlighted in colors.}
    \label{fig:performance}
    \vspace{-0.3cm}
\end{figure*}

\vspace{-0.1cm}
\subsection{Qualitative Studies}
\vspace{-0.1cm}
To demonstrate the open-world capability of our RegionPLC, we provide compelling qualitative results showcasing its capability in recognizing and localizing novel categories. As illustrated in Figure~\ref{fig:performance} (a), RegionPLC successfully identifies numerous categories without any human annotation, demonstrating the quality and richness of our region-level captions and the effectiveness of our region-aware point-discriminative learning objective.
For base-annotated cases, our model can recognize challenging tail classes such ``keyboard''  and ``laddar'' with precise segmentation in indoor scenarios (see Figure~\ref{fig:performance} (b)) and small-scale objects with only a few points such as ``motorcycle'' in outdoor scenarios (see Figure~\ref{fig:performance} (c)). Moreover, RegionPLC shows a strong localization ability in open-world instance segmentation, accurately grouping novel objects as shown in Figure~\ref{fig:performance} (d).

\section{Ablation Study}
\vspace{-0.2cm}
In this section, we examine key components of our framework through in-depth ablation studies. Results for base-annotated and annotation-free experiments are measured in hIoU / mIoU$^\mathcal{B}$ / mIoU$^\mathcal{N}$ and mIoU$^\dag$ (mAcc$^\dag$) separately. 

\begin{table}[h]
    \vspace{-0.3cm}
    \centering
    \begin{small}
    \setlength\tabcolsep{2.5pt}
    \scalebox{0.93}{
        \begin{tabular}{c|c|c|c|c|c|c}
            \bottomrule[1pt]
            \multicolumn{5}{c|}{Components} & {ScanNet} &  ScanNet \\
            \cline{1-5}
            $\mathbf{t}^{v+e}$ & $\mathbf{t}^r$ & $\mathcal{L}_{\text{pdc}}$ & $\mathcal{L}_{\text{rpdc}}$ & SFusion & B0/N17 & B12/N7 \\
            \hline
              & & & & & 0.3 (5.3) & 24.5 / 70.0 / 14.8 \\
              $\checkmark$ & & & & & 17.7 (33.5) & 55.3 / 69.5 / 45.9 \\
              & $\checkmark$ & &  & & 21.7 (37.1) & 62.6 / 69.9 / 56.7 \\ 
              & $\checkmark$ & $\checkmark$ & & & 50.6 (71.1) & 63.6 / \textbf{70.6} / 57.9 \\
              & $\checkmark$ &  & $\checkmark$ & &51.7 (72.7) & 67.4 / 70.5 / 64.6 \\ 
              & $\checkmark$ &  & $\checkmark$ & $\checkmark$ & \textbf{56.9} (\textbf{75.5}) & \textbf{68.2} / 69.9 / \textbf{66.6} \\
            \toprule[0.8pt]
        \end{tabular}
    }
    \end{small}
    \vspace{-0.4cm}
    \caption{Component analysis on ScanNet. $\mathbf{t}^{v+e}$ and $\mathbf{t}^r$ denotes the combination of view and entity language supervision~\cite{ding2022language} and best region-level language supervision, respectively.}
    \vspace{-0.5cm}
    \label{tab:component}
\end{table}

\vspace{0.05in}\noindent\textbf{Component Analysis.}
Here, we study the effectiveness of our proposed regional captions $\mathbf{t}^r$, the SFusion strategy for caption integration, point-discriminative contrastive loss $\mathcal{L}_\text{pdc}$ and its region-aware variants $\mathcal{L}_\text{rpdc}$. As shown in Table~\ref{tab:component}, when compared to view- and entity-level captions used in PLA~\cite{ding2022language}, our region-level language supervision delivers consistent boosts about $4\%\sim10.8\%$ across different category partitions.
Additionally, $\mathcal{L}_{\text{pdc}}$ achieves considerable gains when paired with $\mathbf{t}^r$. Particularly, it brings $28.9\%$ mIoU$^\dag$ gains in the annotation-free setting, illustrating its superiority in learning point-discriminative features for dense parsing tasks. When combined with the region-aware factor, $\mathcal{L}_{\text{rpdc}}$ surpasses $\mathcal{L}_{\text{pdc}}$ by $1.1\%\sim3.8\%$ mIoU$^\dag$. Lastly, $2\%\sim5.2\%$ improvements yielded from SFusion strategy confirm its effectiveness in eliminating redundancy and conflicts from multiple captions for training.

\begin{table}[h!]
\centering
\vspace{-0.3cm}
    \begin{subtable}{0.48\textwidth}
    \centering
    \begin{small}
    \setlength\tabcolsep{3pt}
    \scalebox{0.9}{
        \begin{tabular}{c|c|c|c|c}
            \bottomrule[1pt]
            \multirow{2}{*}{\makecell{Caption source}} & \multicolumn{4}{c}{[$T_{l}$, $T_{h}$]} \\
            \cline{2-5}
             & [0.0, 1.0]$^\ast$ & [0.5, 1.0] & [0.0, 0.5] & [0.0, 0.2] \\
            \hline
            $\mathbf{t}^\text{kos}$, $\mathbf{t}^\text{sw}$ & 53.1 (73.9) & 54.6 (75.3) & 54.3 (74.4) & 56.6 (74.7) \\
            $\mathbf{t}^\text{kos}$, $\mathbf{t}^\text{det-t}$ & 54.0 (74.5) & 54.1 (73.7) & 54.3 (73.9) & 55.9 (76.1) \\
            $\mathbf{t}^\text{kos}$, $\mathbf{t}^\text{sw}$, $\mathbf{t}^\text{det-t}$ & 54.9 (74.2) & 55.2 (73.7) & 55.4 (75.3) & 56.9 (75.5) \\
            \toprule[0.8pt]
        \end{tabular}
    }
    \end{small}
    \vspace{-0.1cm}
    \caption{Ablation of various caption sources and caption overlap thresholds. $^\ast$ equals to the data-mixing baseline. We report mIoU (mAcc) here.}
    \end{subtable}
    
    \begin{subtable}{0.48\textwidth}
    \centering
    \begin{small}
    \setlength\tabcolsep{3pt}
    \scalebox{0.9}{
        \begin{tabular}{l|c|c|c|c}
            \bottomrule[1pt]
            [$T_{l}$, $T_{h}$] & [0.5, 1.0] & [0.0, 0.5] & [0.0, 0.5] & [0.0, 0.2] \\
            Ratio ($\epsilon$) & 0.75$^\oplus$ & 0.34$^\oplus$ & 0.75 & 0.72$^\oplus$ \\
            \hline
            mIoU (mAcc) & 54.6 (75.3) & 54.3 (74.4) & 55.4 (75.6) & 56.6 (74.7) \\
            \toprule[0.8pt]
        \end{tabular}
    }
    \end{small}
    \vspace{-0.1cm}
    \caption{Ablation of caption overlap thresholds and their ratios when fusing $\mathbf{t}^\text{kos}$ and $\mathbf{t}^\text{sw}$. $^\oplus$ means the raw ratio for $\mathbf{t}^\text{kos}$ and $\mathbf{t}^\text{kos}$ + $\mathbf{t}^\text{sw}$ with no $\epsilon$ applied.}
    \end{subtable}
    \vspace{-0.7cm}
    \caption{SFusion results for zero-shot semantic segmentation considering caption sources, overlap thresholds, and ratios.}\label{tab:sfusion}
    \vspace{-0.4cm}
\end{table}

\vspace{0.01cm}
\noindent\textbf{SFusion.} We also study the effectiveness of our SFusion strategy. As shown in Table~\ref{tab:sfusion} (b), with a similar ratio $\epsilon$ of the main caption source relative to all merged captions, fusing 3D-language pairs that have a low spatial overlap ratio (i.e., less than 0.5) yields superior results compared to fusing pairs that are highly overlapped (i.e., greater than 0.5). Besides, as shown in Table~\ref{tab:sfusion} (a), our SFusion largely outperforms the naive data-mixing strategy ([$T_{l}$, $T_{h}$]=[0.0, 1.0]) with $1.9\% \sim 3.5\%$ gains. These experimental results affirm that potential conflicts and redundancies introduced by regions with a high degree of overlap restrict the benefits derived from multiple language sources. Fortunately, our SFusion method effectively tackles this challenge.

\begin{figure}
    \centering
    \includegraphics[width=1\linewidth]{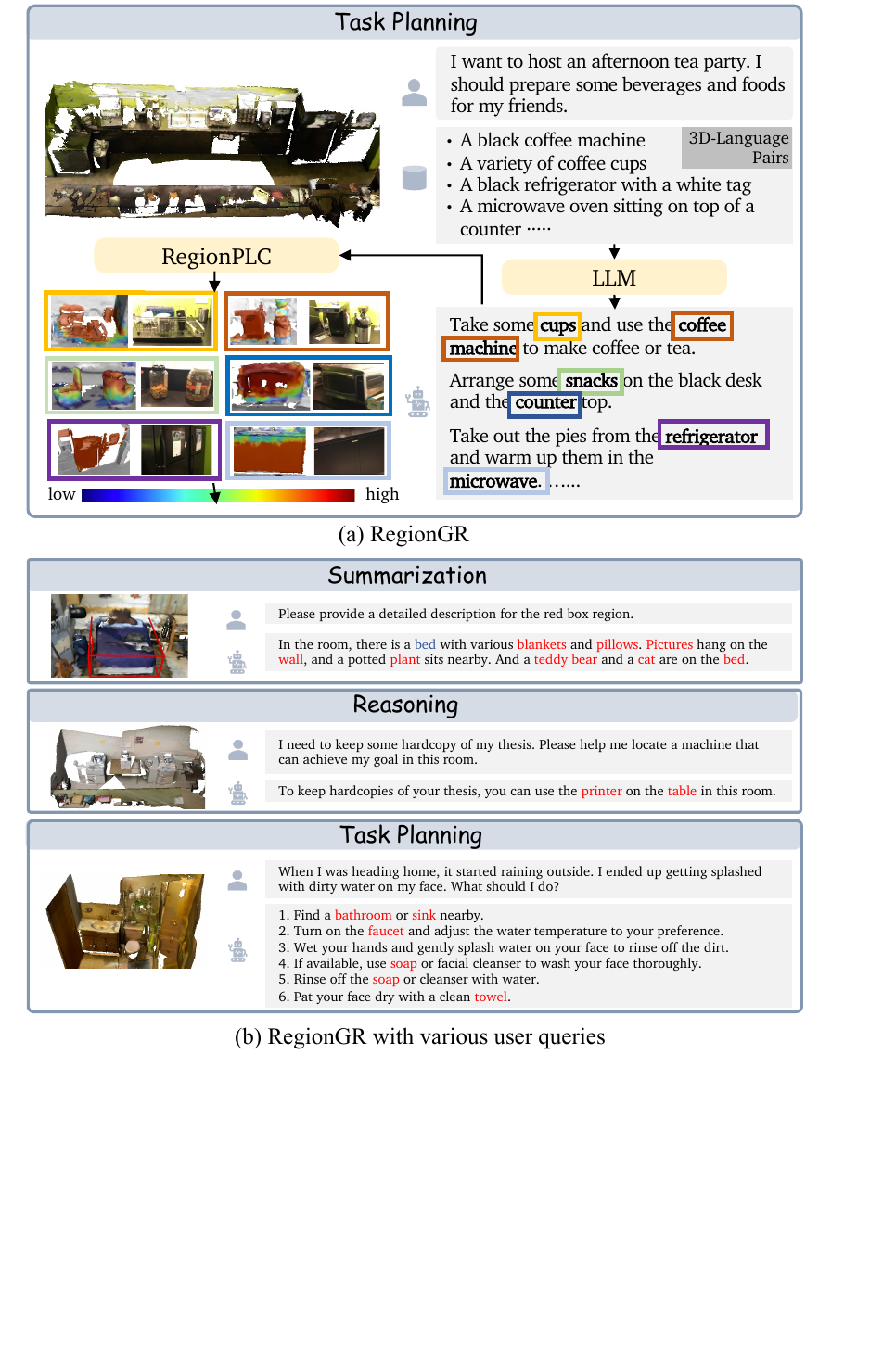}
    \vspace{-0.7cm}
    \caption{(a) Visualizations of RegionGR that integrates LLM for open-ended grounded 3D reasoning. (b) Demonstrating the versatility of RegionGR via more examples of answering user queries.}
    \label{fig:application}
    \vspace{-0.6cm}
\end{figure}

\vspace{-0.2cm}
\section{Open-ended Grounded 3D Reasoning}
\vspace{-0.1cm}
Recently, there has been a growing interest in employing language as an interface for connecting human intentions with visual understanding, which facilitates high-level reasoning and planning in the development of embodied agents. Without specific design, RegionPLC can be seamlessly integrated with large language models to enable open-ended grounded 3D reasoning, referred to as RegionGR. 
As depicted in Figure~\ref{fig:application} (a), RegionGR integrates large language models (LLM) for 3D reasoning with regional 3D-language pairs as a knowledge base and utilizes RegionPLC to coarsely locate and identify corresponding objects within the 3D scene for grounded reasoning. Moreover, Figure~\ref{fig:application} (b) further 
exhibits the versatility of RegionGR in responding to user intentions, from summarization to reasoning and planning, particularly within user-specified regions of interest (as shown in the summarization example). 

Specifically, RegionGR runs in three steps: ($i$) We first initialize ``environment context'' with our \textit{regional 3D-language pairs}, which enables LLM to understand a given scene. If a user specifies an interested 3D region (see Figure~\ref{fig:application} (b) left), highly overlapped 3D-language pairs are kept. ($ii$) We then feed \textit{our prompt in Sec.~S4 of suppl.} along with ``user query'' and ``environment context'' into LLM to generate answers. ($iii$) Finally, we parse objects from LLM's response and ground them with RegionPLC.

\vspace{-0.2cm}
\section{Conclusion}
\vspace{-0.1cm}
We present RegionPLC, a holistic regional point-language contrastive learning framework to recognize and localize unseen categories in open-world 3D scene understanding. By leveraging advanced VL models and our SFusion strategy, RegionPLC effectively builds comprehensive regional point-language pairs. Furthermore, our region-aware point-discriminative contrastive loss aids in learning distinctive and robust features from regional captions. Extensive experiments demonstrate that RegionPLC remarkably outperforms prior open-world methods in both indoor and outdoor scenarios and excels in challenging long-tail or annotation-free scenarios. Besides, RegionPLC can be effortlessly integrated with LLM for grounded 3D visual reasoning.

\vspace{0.1cm}
\noindent\textbf{Acknowledgement} This work has been supported by Hong Kong Research Grant Council - Early Career Scheme (Grant No. 27209621), General Research Fund Scheme (Grant No. 17202422), and RGC Matching Fund Scheme (RMGS). Part of the described research work is conducted in the JC STEM Lab of Robotics for Soft Materials funded by The Hong Kong Jockey Club Charities Trust.

{
    \small
    \bibliographystyle{ieeenat_fullname}
    \bibliography{main}
}

\appendix
\maketitlesupplementary

\centerline{\large{\textbf{Outline}}}
\vspace{0.3cm}
In this supplementary file, we provide more experimental results and details not elaborated in our main paper due to page length limits:
\begin{itemize}
    \item Sec.~\ref{sec:implementation}: Implementation details of RegionPLC and baseline methods.
    \item Sec.~\ref{sec:result}: Additional experimental results on per-class performance, zero-shot domain transfer results, and error-bar analysis. 
    \item Sec.~\ref{sec:vis}: Qualitative results of RegionPLC.
    \item Sec.~\ref{sec:prompts_for_regiongr}: Prompts for LLM in RegionGR.
    \item Sec.~\ref{sec:limitation}: Limitation and open problems.
\end{itemize}

\section{Implementation Details}
\label{sec:implementation}

Here, we present the implementation details of our RegionPLC, 
dataset category partition and the implementation of baseline methods.

\subsection{Implementation Details of RegionPLC}\label{sec:supp_regionplc_implement}
\noindent\textbf{Network Architecture.}
The network architecture of RegionPLC is the same as PLA~\cite{ding2022language} (\ie SparseUNet16), ensuring a fair comparison on ScanNet~\cite{dai2017scannet}. 
For ScanNet200~\cite{rozenberszki2022language}, we augment the base hidden dimension (\ie the hidden dimension of the bottleneck) of sparse UNet from 16 to 32 (\ie SparseUNet32), yielding better performance on this complex, long-tail dataset.
In nuScenes~\cite{caesar2020nuscenes}, we adopt the same backbone used in ScanNet200 with 5 residual blocks and a base hidden dimension of 32.
In addition, in the base-annotated open-world setting, we follow PLA~\cite{ding2022language} to employ the binary encoder with binary loss, classification head with semantic segmentation loss and instance head with instance loss on base categories.

\vspace{0.05in}\noindent\textbf{Training Schedule.}
We train 512 epochs on ScanNet and ScanNet200, and 50 epochs on nuScenes for semantic segmentation, and 640 epochs for ScanNet instance segmentation. The initial learning rate is set as 0.004 for ScanNet and ScanNet200 and 0.01 for nuScenes. The learning rate decays in cosine and polynomials on ScanNet and nuScenes, respectively.

\vspace{0.05in}\noindent\textbf{Regional 3D-Language pairs.}
Regarding vision-language (VL) models that generate captions, we use OFA~\cite{wang2022ofa} for generating $\mathbf{t}^{\text{sw}}$ and $\mathbf{t}^{\text{det-c}}$ as the caption model. 
As for the object proposal in $\mathbf{t}^{\text{det-t}}$ and $\mathbf{t}^{\text{det-c}}$, we use Detic~\cite{zhou2022detecting} with LVIS~\cite{gupta2019lvis} vocabulary space. The prompt template of $\mathbf{t}^{\text{det-t}}$ is the same as CLIP~\cite{radford2021learning}. 
Other VL foundation models are also feasible, and a more robust VL foundation model should enhance our methods' performance through providing higher-quality object proposals and language descriptions. By default, we use 125K frames in the  ScanNet dataset and all images in the nuScenes dataset to extract their regional captions, respectively.

\vspace{0.05in}\noindent\textbf{Inference Cost Analysis.}
In the annotation-free setting of the main paper, we evaluate the efficiency of open-world models in terms of training hours (on 8 NVIDIA A100 GPUs), extra storage usage and inference latency (on a single NVIDIA 2080Ti GPU), since they impose a significant overhead during training or inference. 
The extra storage for RegionPLC and PLA~\cite{ding2022language} is to store 3D-language pairs, while for OpenScene~\cite{peng2022openscene} is to save fused 2D features.

\vspace{0.05in}\noindent\textbf{Category Prompt in nuScenes.}
We use a category prompt for nuScenes to replace ambiguous words within the category names such as ``manmade'' and ``driveable\_surface''. The concrete category mapping is illustrated in Table~\ref{tab:nuscenes_category_prompt}. 

\begin{table}[h!]
    \centering
    \scalebox{0.95}{
    \begin{tabular}{c|l}
    \bottomrule[1pt]
        Category Name & Category Prompts \\
        \hline
        \hline
        barrier  & barrier or fence \\
        \hline
        bicycle & bicycle or bike or cycle \\
        \hline
        bus & bus  \\
        \hline
        car & car \\
        \hline
        construction\_vehicle & \makecell[l]{construction vehicle or bulldozer\\ or  excavator or concrete mixer or\\ crane  or dump truck} \\
        \hline
        motorcycle &  motorcycle or motorbike \\
        \hline
        pedestrian & person or people or man or woman \\
        \hline
        traffic\_cone & traffic cone \\
        \hline
        trailer & trailer \\
        \hline
        truck & truck \\
        \hline
        driveable\_surface & road or street \\
        \hline
        other flat & other flat \\
        \hline
        sidewalk & sidewalk \\
        \hline
        terrain & grass or rolling hills or soil or gravel \\
        \hline
        manmade & \makecell[l]{building or wall or fence or pole \\ or sign or traffic light} \\
        \hline
        vegetation & \makecell[l]{bushes or plants or trees or \\ potted plants}\\
        \toprule[0.8pt]
    \end{tabular}}
    \caption{The category prompt for nuScenes.}
    \label{tab:nuscenes_category_prompt}
\end{table}

\begin{table*}[htbp]
    \centering
    \begin{small}
    \setlength\tabcolsep{5pt}
    \scalebox{1.0}{
        \begin{tabular}{l|l|l}
            \bottomrule[1pt]
            Partition & Base Categories & Novel Categories \\
            \hline
            B12/N3 & \makecell[l]{barrier, bicycle, bus, car, construction\_vehicle, trailer, truck, \\ diveable\_surface, sidewalk, terrain, manmade, vegetation} & traffic\_cone, motorcycle, pedestrian\\
            \hline
            B10/N5 & \makecell[l]{bicycle, bus, car, construction\_vehicle, trailer, truck, \\ driveable\_surface, terrain, manmade, vegetation} & \makecell[l]{barrier, motorcycle, pedestrian, traffic\_cone, sidewalk}\\
            \toprule[0.8pt]
        \end{tabular}
    }
    \end{small}
    \caption{Category partitions for open-world semantic segmentation on nuScenes.}
    \label{tab:nuscene_partition}
\end{table*}

\begin{table*}[htbp]
    \centering
    \begin{small}
    \setlength\tabcolsep{5pt}
    \scalebox{1.0}{
        \begin{tabular}{l|l}
            \bottomrule[1pt]
            Partition &  Novel Categories \\
            \hline
            B170/N30 & \makecell[l]{pillow, box, clothes, counter, dresser, keyboard, backpack, printer, shower curtain, bin, copier, sofa chair, \\ recycling bin, clock, guitar, seat, ladder, cup, toaster, ironing board, toilet seat cover dispenser, furniture, cart, \\ projector, shower floor, laundry detergent, bathroom stall door, dumbbell, folded chair, mattress}\\
            \hline
            B150/N50 & \makecell[l]{couch, window, bookshelf, coffee table, kitchen cabinet, clothes, counter, end table, bag, backpack, printer, \\ microwave, shoe, bin, washing machine, sofa chair, paper, blinds, radiator, recycling bin, soap dispenser,\\ bucket, stand, light, pipe, bathroom stall, cup, storage bin, coffee maker, machine, fireplace, mini fridge, hat,\\ cart, light switch, decoration, plunger, stuffed animal, dish rack, broom, range hood, water pitcher, paper bag, \\ bathroom vanity ceiling light, trash bin, stair rail, coat rack, calendar, poster}\\
            \toprule[0.8pt]
        \end{tabular}
    }
    \end{small}
    \caption{Novel Categories for open-world semantic segmentation on ScanNet200. We only present novel categories here as there are too many base categories to show here. The partition of base categories can be easily obtained by carrying on the set difference between all categories and novel categories.}
    \label{tab:scannet200_partition}
\end{table*}

\subsection{Category Partition for Base-annotated Results}
In the base-annotated open-world setting, we divide all categories into base and novel.
As for ScanNet~\cite{dai2017scannet}, we follow the category partition of PLA~\cite{ding2022language}.
As for nuScenes~\cite{caesar2020nuscenes}, we discard the ambiguous category ``otherflat'' and split the remaining 15 categories as illustrated in Table~\ref{tab:nuscene_partition}. We also randomly split 30 and 50 novel categories for ScanNet200~\cite{rozenberszki2022language}, as shown in Table~\ref{tab:scannet200_partition}. Notably, 11 categories absent from the ScanNet200 validation set are consistently partitioned into base categories (\ie training set) to guarantee sufficient novel categories for validation. These 11 train-only categories in ScanNet200 are ``bicycle'', ``storage container'', ``candle'', ``guitar case'', ``purse'', ``alarm clock'', ``music stand'', ``cd case'', ``structure'', ``storage organizer'' and ``luggage''.

\begin{table*}[htbp]
    \vspace{-0.1cm}
    \centering
    \begin{small}
    \scalebox{0.9}{
        \begin{tabular}{l|c|c|c|c|c|c}
            \bottomrule[1pt]
            Method & 2D models & \makecell{Multi-view \\inference} &  Learning in 3D & \makecell{Scale up with\\ better 3D backbone} & Extra storage & Supervision \\
            \hline
            ConceptFusion~\cite{jatavallabhula2023conceptfusion} & \makecell{Mask2Former~\cite{cheng2021mask2former} \& \\ CLIP~\cite{radford2021learning}} &  $\checkmark$ & $\times$ & $\times$ & $\times$ & $\times$  \\
            \hline
            OpenScene-3D~\cite{peng2022openscene} & \makecell{LSeg~\cite{li2022languagedriven} \& \\ OpenSeg~\cite{Ghiasi2021ScalingOI}} & $\times$ & $\checkmark$ & $\times$ & high & Pixel-aligned 2D features  \\
            \hline
            PLA~\cite{ding2022language} & VIT-GPT2~\cite{vit-gpt2} & $\times$ & $\checkmark$ & $\checkmark$ & low & Sparse language supervision \\
            \hline\hline
            RegionPLC & \makecell{OFA~\cite{wang2022ofa} \& \\ Detic~\cite{zhou2022detecting} \& \\ Kosmos-2~\cite{peng2023kosmos2}} & $\times$ & $\checkmark$ & $\checkmark$ & low & Dense language supervision  \\
            \toprule[0.8pt]
        \end{tabular}
     }
    \end{small}
    \vspace{-0.3cm}
    \caption{Comparison between different 3D open-world scene understanding methods.}
    \label{tab:compared_methods}
    \vspace{-0.2cm}
\end{table*}

\begin{table*}[htbp]
    \centering
    \begin{small}
      \scalebox{0.92}{
        \setlength{\tabcolsep}{1.3mm}{
        \begin{tabular}{c|c|ccccccccccccccccccc}
            \bottomrule[1pt]
            Method & Partition & \rotatebox[origin=c]{90}{wall} & \rotatebox[origin=c]{90}{floor} & \rotatebox[origin=c]{90}{cabinet} & \rotatebox[origin=c]{90}{bed} & \rotatebox[origin=c]{90}{chair} & \rotatebox[origin=c]{90}{sofa} & \rotatebox[origin=c]{90}{table} & \rotatebox[origin=c]{90}{door} & \rotatebox[origin=c]{90}{window} & \rotatebox[origin=c]{90}{ bookshelf } & \rotatebox[origin=c]{90}{picture} & \rotatebox[origin=c]{90}{counter} & \rotatebox[origin=c]{90}{desk} & \rotatebox[origin=c]{90}{curtain} & \rotatebox[origin=c]{90}{fridge} & \rotatebox[origin=c]{90}{shower c.} & \rotatebox[origin=c]{90}{toilet} & \rotatebox[origin=c]{90}{sink} & \rotatebox[origin=c]{90}{bathtub} \\
            \hline
            \multirow{3}{*}{PLA~\cite{ding2022language}} & B15/N4 & 84.6 & 95.0 & 64.9 & 81.1 & 87.9 & {\cellcolor{myblue2}75.9} & 72.2 & 61.9 & 62.1 & {\cellcolor{myblue2}69.5} & 30.9 & 60.1 & {\cellcolor{myblue2}46.5} & 70.7 & 50.5 & 66.1 & {\cellcolor{myblue2}56.8} & 59.0 & 81.7 \\
            & B12/N7 & 84.7 & 95.1 & 65.3 & {\cellcolor{myblue2}57.8} & {\cellcolor{myblue2}44.2} & 75.9 & {\cellcolor{myblue2}34.5} & 62.5 & 62.3 & {\cellcolor{myblue2}62.1} & {\cellcolor{myblue2}20.5} & 57.8 & 61.4 & 72.4 & 47.9 & 64.9 & 85.9 & {\cellcolor{myblue2}28.4} & {\cellcolor{myblue2}69.6} \\
            & B10/N9 & 83.8 & 95.2 & 64.3 & 80.9 & 88.0 & 78.5 & 73.2 & 60.6 & 61.5 & {\cellcolor{myblue2}68.6} & {\cellcolor{myblue2}17.7} & {\cellcolor{myblue2}23.4} & {\cellcolor{myblue2}51.3} & 70.6 & {\cellcolor{myblue2}25.7} & {\cellcolor{myblue2}38.2} & {\cellcolor{myblue2}51.3} & {\cellcolor{myblue2}27.3} & {\cellcolor{myblue2}61.7} \\
            \hline
            \multirow{3}{*}{RegionPLC} & B15/N4 & 84.2 & 95.1 & 66.6 & 81.2 & 88.2 & {\cellcolor{myblue2}81.3} & 72.6 & 61.4 & 60.7 & {\cellcolor{myblue2}75.3} & 30.4 & 57.7 & {\cellcolor{myblue2}53.4} & 70.6 & 46.1 & 64.6 & {\cellcolor{myblue2}72.6} & 59.4 & 84.0 \\
            & B12/N7 & 84.9 & 95.1 & 65.2 & {\cellcolor{myblue2}76.3} & {\cellcolor{myblue2}79.5} & 75.8 & {\cellcolor{myblue2}64.3} & 60.0 & 64.3 & {\cellcolor{myblue2}77.9} & {\cellcolor{myblue2}31.1} & 56.7 & 65.7 & 72.7 & 49.5 & 65.6 & 83.4 & {\cellcolor{myblue2}55.5} & {\cellcolor{myblue2}81.9} \\
            & B10/N9 & 84.3 & 95.2 & 65.5 & 80.6 & 89.2 & 82.7 & 73.8 & 59.6 & 62.0 & {\cellcolor{myblue2}79.7} & {\cellcolor{myblue2}25.0} & {\cellcolor{myblue2}47.7} & {\cellcolor{myblue2}56.3} & 69.8 & {\cellcolor{myblue2}38.0} & {\cellcolor{myblue2}53.2} & {\cellcolor{myblue2}74.4} & {\cellcolor{myblue2}46.6} & {\cellcolor{myblue2}78.9} \\
            \toprule[1pt]
        \end{tabular}}}
    \end{small}
    \caption{Per-class results of base-annotated open-world 3D semantic segmentation on ScanNet in terms of IoU. Performance on novel categories is marked in \colorbox{myblue2}{blue}.}
    \label{tab:per_class_scannet}
\end{table*}

\subsection{Implementation of Baseline Methods}
We re-produce the baseline methods including MaskCLIP~\cite{zhou2022maskclip}, PointCLIP-Seg~\cite{zhang2022pointclip} and OpenScene~\cite{peng2022openscene} for annotation-free open-world semantic segmentation in ScanNet. Details are as follows.

\vspace{0.05in}\noindent\textbf{PointCLIP-Seg and MaskCLIP.}
To apply MaskCLIP~\cite{zhou2022maskclip} on 3D segmentation, we assemble its predictions on multi-view images and back-project them to 3D space as \cite{ding2022language}.
As for PointCLIP~\cite{zhang2022pointclip}, it cannot be directly utilized for the semantic segmentation task, so we extend a segmentation version by modifying the attentive pooling layer of CLIP~\cite{radford2021learning}, as per the method used in MaskCLIP~\cite{zhou2022maskclip}. It is named as PointCLIP-Seg. The major distinction between PointCLIP-Seg and MaskCLIP lies in that PointCLIP-Seg uses depth images rather than RGB images for extracting 2D features.

\vspace{0.05in}\noindent\textbf{OpenScene.} We use the official fused feature released by OpenScene~\cite{peng2022openscene} and its prompt engineering techniques to obtain OpenScene-2D results. 
To ensure a fair comparison, we train OpenScene-3D using the same training schedule and 3D backbone as our RegionPLC. This allows us to compare performance under the same conditions and analyze the results more accurately.

\vspace{0.05in}\noindent\textbf{PLA.} 
As for PLA~\cite{ding2022language} in the annotation-free open-world setting, we only carry on the point-language contrastive learning and discard its binary encoder as there is no annotated base category in the training set.

\subsection{Comparisons of 3D Open-world Scene Understanding Methods}
As shown in Table~\ref{tab:compared_methods},  we compare our RegionPLC to other three cutting-edge 3D open-world scene understanding methods: ConceptFusion~\cite{jatavallabhula2023conceptfusion}, OpenScene~\cite{peng2022openscene} and PLA~\cite{ding2022language}. 
ConceptFusion~\cite{jatavallabhula2023conceptfusion} relies on a multi-view fusion of image predictions during its inference phase. However, its inability to learn from 3D point clouds makes it difficult to extract 3D geometric information.
On the other hand, OpenScene-3D~\cite{peng2022openscene} can learn directly from the 3D point cloud, but this approach necessitates significant additional storage for saving fused 2D features, making it unsuitable for handling large-scale datasets. Furthermore, the ceiling of its performance is limited by the 2D semantic feature and distillation strategy, making it harder to integrate with more advanced 3D backbones.
PLA~\cite{ding2022language}, while only requiring minimal additional storage and being scalable due to only 3D-language supervisions during training, is restricted in its performance by the sparseness and roughness of its language supervisions.
In contrast, our RegionPLC inherits all the strengths of PLA~\cite{ding2022language} and incorporates more advanced 2D models for regional 3D-language association, thereby boosting its open-world capability.

\section{More Experimental Results}
\label{sec:result}
In this section, we present some supplementary experimental results, in addition to the ones provided in our main paper. This part consists of a detailed analysis of the per-class performance, an error-bar analysis and the zero-shot domain transfer experiments.

\subsection{Per-category Results}
Here, we show the per-category performance comparison between PLA~\cite{ding2022language} and RegionPLC for base-annotated open-world 3D semantic segmentation on ScanNet~\cite{dai2017scannet}. As shown in Table~\ref{tab:per_class_scannet}, our RegionPLC obtains improvements on all novel categories across different partitions, which demonstrates its effectiveness.

\begin{table*}[htbp]
    \vspace{-0.1cm}
    \centering
    \begin{small}
    \setlength\tabcolsep{8pt}
    \scalebox{1.0}{
        \begin{tabular}{l|c|c|c|c}
            \bottomrule[1pt]
            \multirow{2}{*}{Round}  & \multicolumn{3}{c|}{Base-annotated ScanNet~\cite{dai2017scannet}} & \multirow{2}{*}{Annotation-free ScanNet~\cite{dai2017scannet}} \\
            \cline{2-4}
            & {B15/N4} & {B12/N7} & {B10/N9} & \\
            \hline
            1 & 69.4 / 68.2 / 70.7 & 68.2 / 69.9 / 66.6 & 64.3 / 76.3 / 55.6 & 59.6 (77.5)\\
            \hline
            2 & 69.5 / 68.6 / 70.4 & 67.6 / 69.8 / 65.4 & 63.9 /  76.4 / 54.9 & 59.2 (78.0) \\
            \hline
            3 & 69.7 / 68.6 / 70.8 & 67.7 / 69.4 / 66.1 & 63.8 / 76.1 / 54.9 & 59.1 (76.6) \\
            \toprule[0.8pt]
        \end{tabular}
     }
    \end{small}
    \caption{Repeated results for base-annotated and annotation-free open-world 3D semantic segmentation on ScanNet. Base-annotated results are measured in hIoU / mIoU$^\mathcal{B}$ / mIoU$^\mathcal{N}$, while annotation-free experiments are measured in mIoU$^\dag$ (mAcc$^\dag$).}
    \label{tab:error_bar_base_annotated}
    \vspace{-0.2cm}
\end{table*}

\subsection{Error Bar}
Here, we provide an error bar for our open-world 3D scene understanding framework on both base-annotated and annotation-free settings by reproducing each experiment 3 times. As shown in Table~\ref{tab:error_bar_base_annotated}, the performance of RegionPLC is generally stable on ScanNet open-world segmentation, demonstrating its robustness.

\subsection{Zero-shot Domain Transfer}
We study the zero-shot domain generalization capability of open-world methods by transferring the ScanNet-trained model to S3DIS without fine-tuning.
As shown in Table~\ref{tab:transfer}, RegionPLC enjoys $6.8\% \sim 37.1\%$ boosts compared to PLA~\cite{ding2022language} in mIoU$^\dag$ on different splits. Notice that more base categories on ScanNet can hinder the generalization on S3DIS, indicating that dataset-specific annotation penalizes the model's transferability. In contrast, solely learning from semantic-rich caption supervision achieves great out-of-domain generalization ability.

\begin{table}[htbp]

    \centering
    \begin{small}
    \setlength\tabcolsep{8pt}
    \scalebox{1.0}{
        \begin{tabular}{c|c|c|c}
            \bottomrule[1pt]
            \multirow{2}{*}{\makecell[c]{ScanNet \\ partition}} & \multicolumn{3}{c}{S3DIS Semantic Segmentation} \\
            \cline{2-4}
            & OVSeg-3D~\cite{ding2022language} & PLA~\cite{ding2022language} & RegionPLC \\
            \hline
            B15/N4 & 31.1 (46.6) & 39.1 (56.2) & \textbf{52.2} (\textbf{64.5}) \\
            \hline
            B12/N7 & 23.6 (42.7) & 35.4 (60.4) &  \textbf{45.0} (\textbf{61.5}) \\
            \hline
            B10/N9 & 36.0 (50.9) & 43.7 (60.4) & \textbf{50.5} (\textbf{63.2}) \\
            \hline
            B0/N17 & 01.7 (11.2) & 13.4 (25.1) & \textbf{50.5} (\textbf{67.6}) \\
            \toprule[0.8pt]
        \end{tabular}
    }
    \end{small}
    \caption{Zero-shot domain transfer results for semantic segmentation in items of mIoU$^\dag$ (mAcc$^\dag$) on ScanNet $\rightarrow$ S3DIS.}
    \vspace{-0.4cm}
    \label{tab:transfer}
\end{table}

\section{Qualitative Results for Annotation-free Open World}\label{sec:vis}
\begin{figure*}
    \centering
    \includegraphics[width=0.9\linewidth]{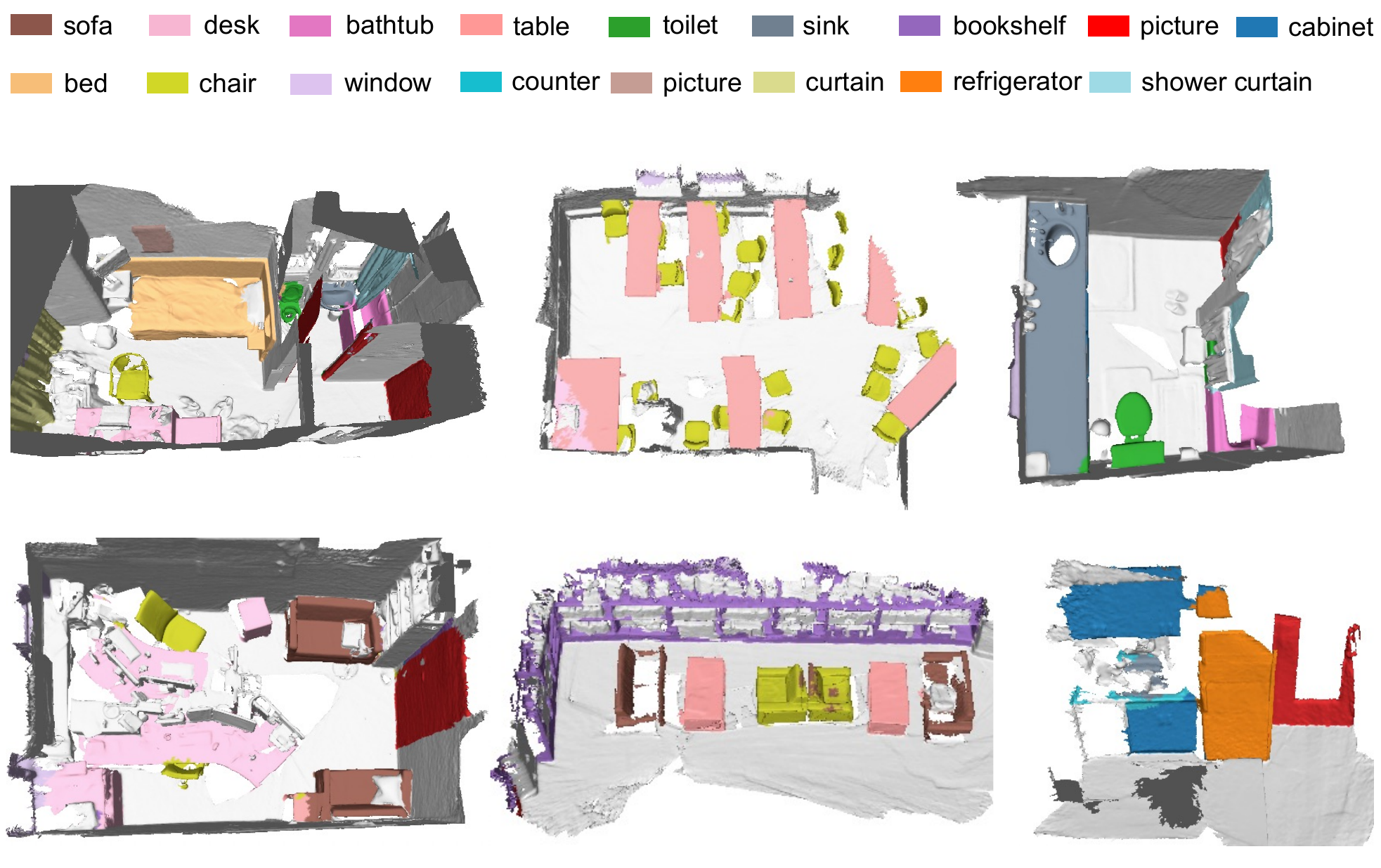}
    \caption{Qualitative results of annotation-free semantic segmentation on ScanNet.} 

    \label{fig:zs_vis}
\end{figure*}

Here, we provide more qualitative results of RegionPLC in the most challenging annotation-free open-world scenario. As shown in Figure~\ref{fig:zs_vis}, our RegionPLC can distinguish different semantics with remarkable segmentation results covering a wide range of categories.

\begin{figure*}
    \centering
    \includegraphics[width=0.9\linewidth]{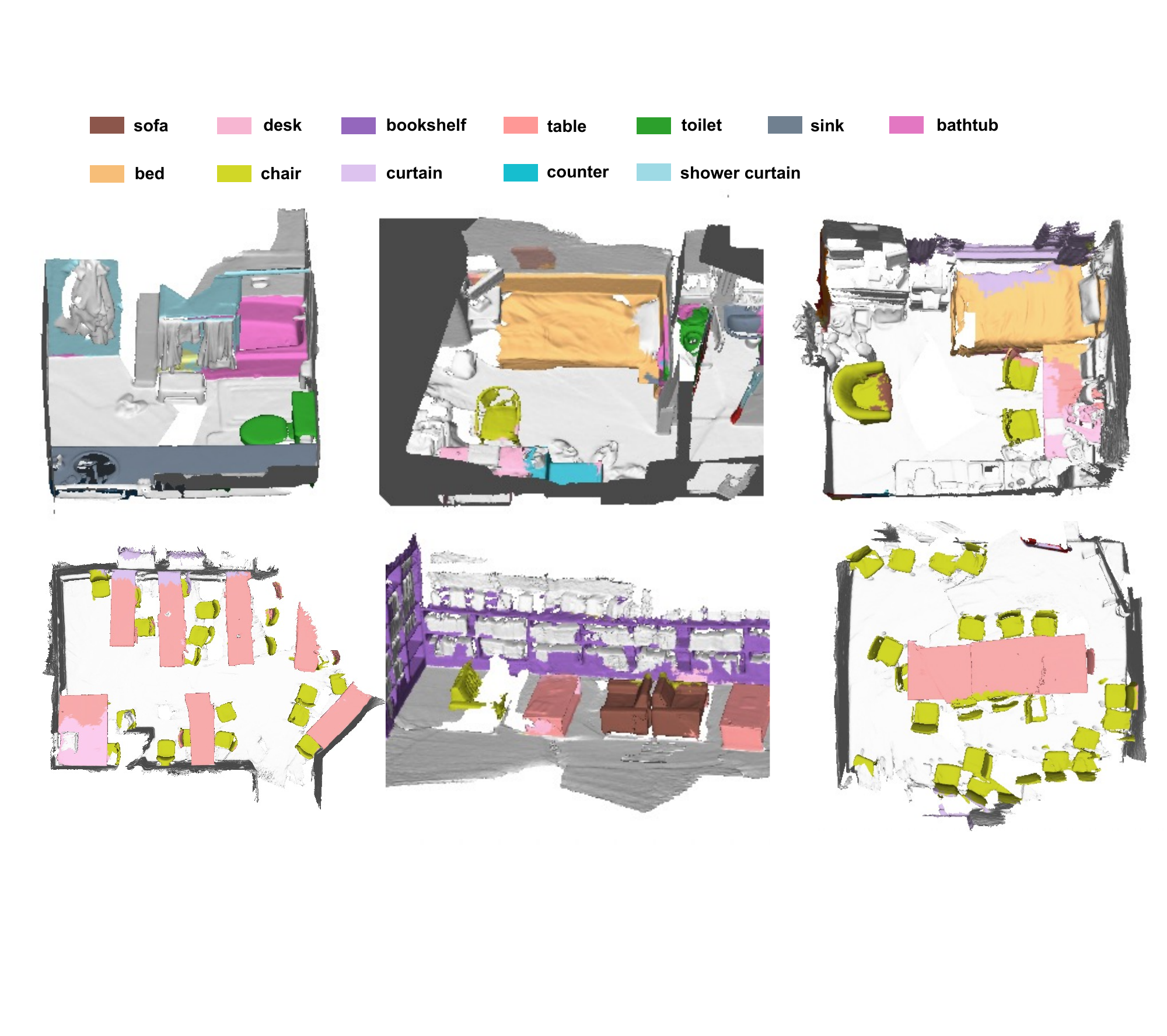}

    \caption{Visualization in heat map of tail classes with the annotation-free model on ScanNet.} 
    \vspace{-0.4cm}
    \label{fig:zs_heatmap_vis}
\end{figure*}

On the other hand, we also explore the potential of our RegionPLC to discover tail and rare categories in real-world scenarios. As shown in Figure~\ref{fig:zs_heatmap_vis}, we visualize the heat maps of the point-wise response given a text query. Our RegionPLC can discover a lot of tail categories such as ``trash can'', ``shoe'' and ``nightstand'' without any human annotation. These results demonstrate the effectiveness of our regional point-language contrastive learning framework in solving open-world 3D scene understanding problems.

\section{Prompts for RegionGR}
\label{sec:prompts_for_regiongr}
As highlighted in the main paper, our RegionPLC is capable of incorporating large language models (LLM), such as GPT-3.5~\cite{schulman2022chatgpt}, to execute grounded 3D reasoning, a pipeline we refer to as RegionGR. The LLM is given human queries and regional captions for the purpose of reasoning.  Note that if a human query pertains to a particular 3D region, we will filter captions, retaining only those that show significant overlap with the specified 3D region as the input. The prompt example we used is as follows.

\begin{quote}
{\small  
\texttt{[Role]} \\
\texttt{You are a household manager. Your job is to understand human instructions, and you should give step-by-step suggestions according to the provided environmental context.}\\

\texttt{[Task]}\\
\texttt{Your task is to give a suitable response to the <question> according to the <env\_context>; if possible, respond in detail with clear logic. Both the question and env context are given, delimited by triple quotes.}\\

\texttt{[Env Context]}\\
\texttt{Here is the env context, containing some words, phrases, or short sentences describing contents in a 3D room. Answer the user's request based on this.}\\
\texttt{<\{\textbf{env}\}>}\\

\texttt{[Rules]}\\
\texttt{Return answers closely related to the provided information, especially the objects mentioned in the provided context.
Keep the final answer simple and short, within 30 words.
Use natural language like humans do in daily life.}\\

\texttt{[Steps]}\\
\texttt{According to the query, understand the intention behind "What do I want/need to do?"
Find the objects related to my question from the env context.
To give the final answer, you should tell me the operation I need to do and the object I need to interact with.
The answer needs to be realistic, and the objects in your answer need to be based on the provided env context.}\\

\texttt{[Dialog Style]}}\\
\texttt{You should respond in a polite, kind, and natural language tone. Try to talk like a human, but keep it short.}\\

\texttt{Begin Task}\\\\
\texttt{The question: <\{\textbf{question}\}>}\\
\end{quote}

\section{Limitation and Future Works}\label{sec:limitation}
Although our RegionPLC has yielded impressive results in 3D open-world scene understanding with a broad spectrum of unseen categories, certain limitations and potential avenues for enhancement remain. On the one hand, the promising results obtained by the combination of RegionPLC and OpenScene~\cite{peng2022openscene} demonstrate the strong potential to introduce 2D image features as auxiliary supervision for training RegionPLC. The current loss combination is straightforward, and we believe that more advanced combination strategies that integrate language, 3D and image features can lead to better performance.

Another aspect warranting improvement is our utilization of visual prompts, which are pre-defined prior to training and remain unchanged throughout the process. 
Better and more adaptive visual prompting techniques might improve the quality of language supervision.
Moving forward, we are interested in further developing an open-world 3D scene understanding framework that addresses these two limitations.

\end{document}